\documentclass[lettersize,journal]{IEEEtran}

\usepackage{amssymb}
\usepackage{graphicx}
\usepackage{amsfonts}
\usepackage{balance}
\usepackage{booktabs}
\usepackage{amsmath}
\usepackage{tabularx}
\usepackage{subfigure}
\usepackage{multirow}
\usepackage{rotating}
\usepackage{array}
\usepackage{color}

\usepackage{ulem}
\usepackage[ruled]{algorithm2e}

\ifCLASSOPTIONcompsoc
  \usepackage[nocompress]{cite}
\else
  \usepackage{cite}
\fi

\begin{document}
	
\title{DeepCluE: Enhanced Deep Clustering via Multi-layer Ensembles in Neural Networks}

\author{Dong Huang,~\IEEEmembership{Member,~IEEE, }
	Ding-Hua Chen,
	Xiangji Chen,
	Chang-Dong Wang,~\IEEEmembership{Senior Member,~IEEE, }\\
	and~Jian-Huang Lai,~\IEEEmembership{Senior Member,~IEEE, }
	\IEEEcompsocitemizethanks{\IEEEcompsocthanksitem D. Huang, D.-H. Chen, and X. Chen are with the College of Mathematics and Informatics, South China Agricultural University, Guangzhou, China. \protect\\
		E-mail: huangdonghere@gmail.com, dinghuachen@hotmail.com, \protect\\
		xiangjichen@hotmail.com.
		\IEEEcompsocthanksitem C.-D. Wang and J.-H. Lai are with the School of Computer Science and Engineering,
		Sun Yat-sen University, Guangzhou, China.\protect\\
		E-mail: changdongwang@hotmail.com, stsljh@mail.sysu.edu.cn.}
	\thanks{© 2023 IEEE.  Personal use of this material is permitted.  Permission from IEEE must be obtained for all other uses, in any current or future media, including reprinting/republishing this material for advertising or promotional purposes, creating new collective works, for resale or redistribution to servers or lists, or reuse of any copyrighted component of this work in other works.}
}

\maketitle

\begin{abstract}
Deep clustering has recently emerged as a promising technique for complex data clustering. Despite the considerable progress, previous deep clustering works mostly build or learn the final clustering by only utilizing a single layer of representation, e.g., by performing the $K$-means clustering on the last fully-connected layer or by associating some clustering loss to a specific layer, which neglect the possibilities of jointly leveraging multi-layer representations for enhancing the deep clustering performance. In view of this,  this paper presents a \textit{Deep} \textit{Clu}stering via \textit{E}nsembles (DeepCluE) approach, which bridges the gap between deep clustering and ensemble clustering by harnessing the power of multiple layers in deep neural networks. In particular, we utilize a weight-sharing convolutional neural network as the backbone, which is trained with both the instance-level contrastive learning (via an instance projector) and the cluster-level contrastive learning (via a cluster projector) in an unsupervised manner. Thereafter, multiple layers of feature representations are extracted from the trained network, upon which the ensemble clustering process is further conducted. Specifically, a set of diversified base clusterings are generated from the multi-layer representations via a highly efficient clusterer. Then the reliability of clusters in multiple base clusterings is automatically estimated by exploiting an entropy-based criterion, based on which the set of base clusterings are re-formulated into a weighted-cluster bipartite graph. By partitioning this bipartite graph via transfer cut, the final consensus clustering can be obtained. Experimental results on six image datasets confirm the advantages of DeepCluE over the state-of-the-art deep clustering approaches.
\end{abstract}


\section{Introduction}\label{sec:introduction}
\IEEEPARstart{D}{ata}  clustering is a fundamental yet still challenging problem in machine learning and computational intelligence, which aims to partition a set of data samples into a certain number of homogeneous groups (i.e., clusters) \cite{jain2010data}. Traditional clustering algorithms mostly rely on hand-crafted features according to some domain-specific knowledge. However, when faced with high-dimensional complex data, such as images and videos, the traditional clustering algorithms \cite{jain2010data} may lead to sub-optimal clustering results due to the lack of the ability of feature representation learning.

In recent years, the deep learning has gained significant attention with its superior capability of feature representation learning, which provides an effective tool for the clustering analysis of very complex data. Many clustering methods based on deep neural networks, referred to as \emph{deep clustering} methods, have been developed. These existing deep clustering methods can mainly be divided into two categories, namely, the single-stage methods \cite{yang2017towards,xie2016unsupervised,yang2016joint,guo2017improved,ghasedi2017deep,caron2018deep,ji2019invariant,Choudhury2021,huang2020deep,li2021contrastive}  and the two-stage methods \cite{van2020scan,dang2021nearest}. Specifically, the single-stage deep clustering methods seek to jointly learn feature representations and cluster assignments in an end-to-end framework. For example, Xie et al. \cite{xie2016unsupervised} proposed the Deep Embedding Clustering (DEC) method, which aims to learn a mapping from the data space to a lower-dimensional feature space in which it iteratively optimizes a clustering objective with the Kullback-Leibler (KL) divergence loss.
Ji et al. \cite{ji2019invariant} presented the Invariant Information Clustering (IIC) method which learns a clustering function by maximizing the mutual information between the cluster assignments of data pairs. Besides these single-stage methods \cite{yang2017towards,xie2016unsupervised,yang2016joint,guo2017improved,ghasedi2017deep,caron2018deep,ji2019invariant,Choudhury2021,huang2020deep,li2021contrastive}, some recent efforts in designing two-stage deep clustering methods have also been made \cite{van2020scan,dang2021nearest}. Van Gansbeke et al. \cite{van2020scan} proposed the Semantic Clustering by Adopting Nearest neighbors (SCAN) method, which utilizes a pretext task of contrastive learning to mine the nearest neighbors in the first stage, and  performs a further learning and clustering optimization based on the nearest neighbors in the next stage.

\begin{figure*}[!t]
	\begin{center}
		{\subfigure{\includegraphics[width=1.8\columnwidth]{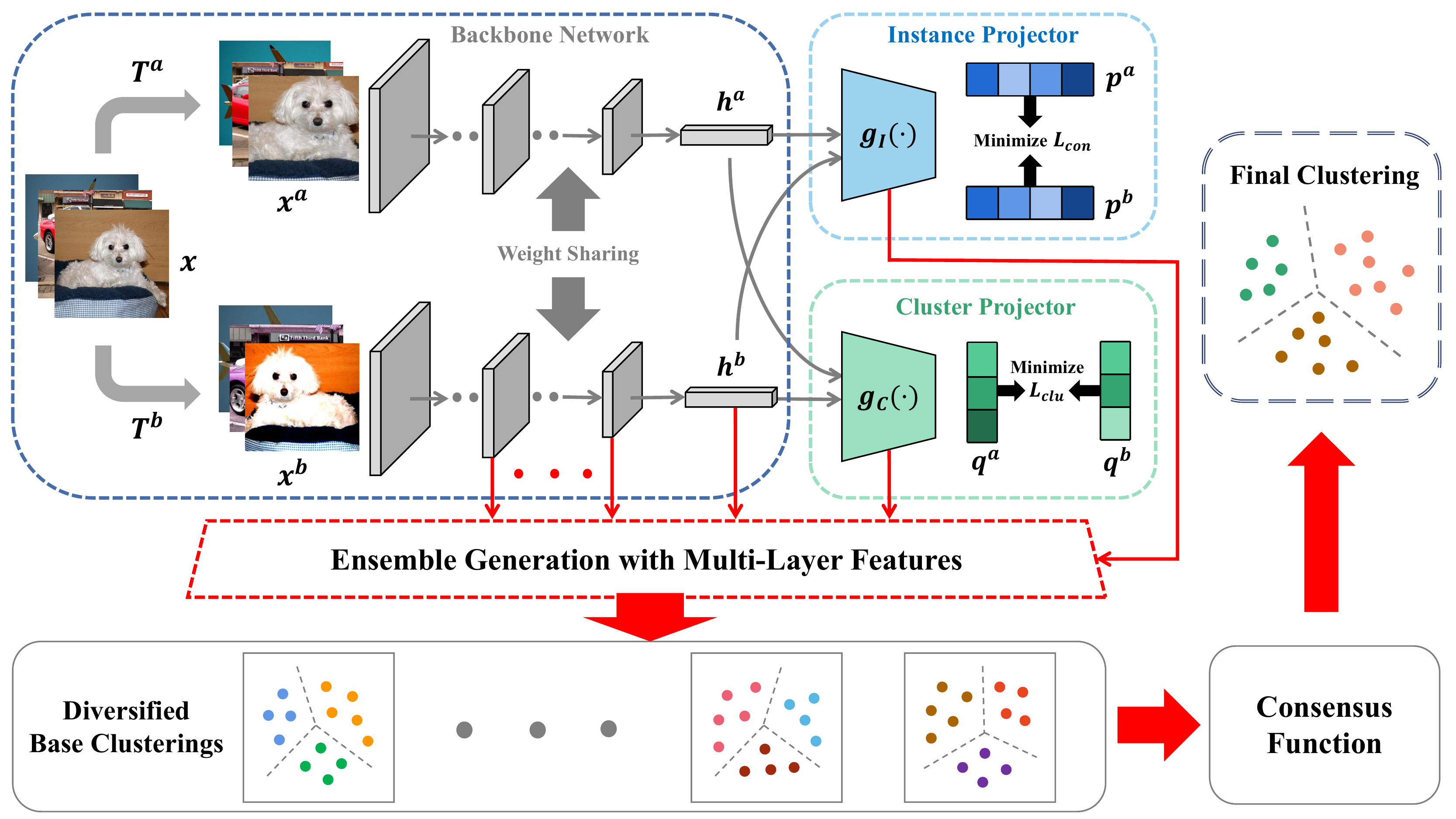}}}
		\caption{An overview of the DeepCluE framework, which first trains the unsupervised contrastive network with the augmented pairs, then generates an ensemble of diversified base clusterings from the output of multiple layers in the network, and finally produce the clustering result via the weighted-cluser bipartite graph based consensus function.}
		\label{fig:architecture}
	\end{center}
\end{figure*}

Though these deep clustering methods \cite{yang2017towards,xie2016unsupervised,yang2016joint,guo2017improved,ghasedi2017deep,caron2018deep,ji2019invariant,Choudhury2021,huang2020deep,li2021contrastive,van2020scan,dang2021nearest,Deng2023,xu19_ijcai,xu22_tcyb} have achieved significant progress in unsupervised representation learning and clustering, they mostly seek to achieve the final clustering by utilizing the feature representation of a single layer (typically the last fully-connected layer) in the neural network. While different layers in a deep neural network can reflect different levels of semantic information and are able to jointly provide a more comprehensive view on the data samples, it is surprising that the previous deep clustering methods mostly overlook the possibilities of jointly leveraging the diverse information of multiple network layers for enhancing the clustering performance. More recently, Li et al. \cite{li2021contrastive} developed the Contrastive Clustering (CC) method by incorporating two types of projectors (i.e., the instance projector and cluster projector) to optimize the instance-level and cluster-level contrastiveness, respectively. However, the instance projector in CC only assists the optimization of the backbone, which does not participate in the clustering process of the cluster projector. More specifically, the CC method still only uses the last layer of the cluster projector for the final clustering, lacking the ability of exploiting the feature information of other modules (or other layers) in the neural network during its clustering process. In spite of these recent progress, it remains an open problem how to jointly exploit the features learned in multiple network layers for enhancing the clustering performance in a unified deep clustering framework.

To address the above problem, in this paper, we present a \textbf{Deep Clu}stering via \textbf{E}nsembles (DeepCluE) approach for unsupervised image clustering, which bridges the gap between deep clustering and ensemble clustering \cite{huang2015robust,yu18_tkde,huang19_tkde,huang23_tkde} and is able to jointly exploit the multi-layer information in deep neural networks (as shown in Fig.~\ref{fig:architecture}). Different from the previous deep clustering approaches that only use a specific layer (typically the last fully-connected layer) in the network for generating the clustering result, our DeepCluE approach for the first time, to the best of our knowledge, leverages the feature representations of multiple network layers for deep image clustering. Specifically, we utilize a weight-sharing convolutional neural network as the backbone to learn the representations of the sample pairs constructed by different data augmentations. Then, two separate projectors, i.e., the instance projector and the cluster projector, are  exploited to enforce the instance-level contrastive learning and the cluster-level contrastive learning, respectively. Further, we simultaneously leverage multiple layers of representations extracted from three modules, i.e., the backbone, the instance projector, and the cluster projector, for the later ensemble clustering process. With consideration to the very different dimensions of multiple network layers, we utilize the principle component analysis (PCA) to reduce the dimension of the output of each convolutional layer, and generate a  set of diversified base clusterings by efficient bipartite graph formulating and partitioning. Thereafter, an entropy-based criterion is exploited to evaluate the reliability of the clusters in different base clusterings, based on which a weighted-cluster consensus function is devised to achieve the clustering result. We conduct experiments on six image datasets, which demonstrate the superiority of DeepCluE over the state-of-the-art.

For clarity, the main contributions of this work are summarized as follows:

\begin{itemize}
	\item This paper bridges the gap between deep clustering and ensemble clustering, and for the first time, to our knowledge, simultaneously leverages the feature representations in multiple network layers for unified deep clustering. Remarkably, our ensemble strategy can serve as an add-on module for any deep clustering models in order to enhance their clustering robustness.
	\item This paper presents a novel deep image clustering approach termed DeepCluE, where the instance-level contrastiveness, the cluster-level contrastiveness, and the ensemble clustering via multi-layer collaboration are integrated into a unified framework.
	\item Extensive experiments are carried out on six image datasets, which confirm (i) the substantial improvement brought in by the multi-layer representations and (ii) the superior clustering performance of DeepCluE over the state-of-the-art deep clustering approaches.
\end{itemize}

The rest of the paper is organized as follows. Section~\ref{sec:related_work} introduces the related works on deep clustering. Section~\ref{sec:proposed_approach} describes the overall process of the proposed DeepCluE approach. The experimental results are reported in Section~\ref{sec:experiments}. Finally, the paper is concluded in Section~\ref{sec:conclusion}.

\section{Related Work}
\label{sec:related_work}

In this paper, we propose a novel deep image clustering approach termed DeepCluE, where the instance-level and cluster-level contrastive learning modules as well as the ensemble clustering module via the joint modeling of multiple network layers are incorporated for enhancing the deep clustering performance. In this section, a literature review on topic of deep clustering will be provided.

Traditional clustering algorithms \cite{cai2009locality,ng2002spectral,yang22_tetci,Cai2023,fang23_tetci} are often designed for low-dimensional vector-like data, which may not perform well on complex high-dimensional data like images. Deep clustering has recently emerged as a promising technique that is able to harness the representation learning power of deep neural networks to transform complex data into some low-dimensional feature representation, upon which some clustering objective can be incorporated to generate the final clustering \cite{yang2017towards,xie2016unsupervised,yang2016joint,guo2017improved,ghasedi2017deep,caron2018deep,ji2019invariant,huang2020deep,li2021contrastive,Choudhury2021,van2020scan,dang2021nearest,Deng2023}.

As one of the earliest studies on this topic, Yang et al. \cite{yang2017towards} proposed the Deep Clustering Network (DCN)  method which  performs the $K$-means clustering on the latent features produced by an auto-encoder, where a reconstruction loss and a clustering loss are jointly minimized. Xie et al. \cite{xie2016unsupervised} utilized a pretrained auto-encoder and iteratively refined the clusters by taking into account their high-confidence assignments with a KL-divergence based clustering loss.
Yang et al. \cite{yang2016joint} presented the JULE method by combining the hierarchical AC process with the deep learning by a recurrent framework. Guo et al. \cite{guo2017improved} developed the Improved Deep Embedded Clustering (IDEC) method that jointly optimizes the cluster label assignments and the feature representation with the local structure of data distribution considered. Dizaji et al. \cite{ghasedi2017deep} incorporated a relative cross-entropy loss and a regularization term (that considers the size of each cluster depending on some prior knowledge) into deep clustering. Caron et al. \cite{caron2018deep} iteratively performed the $K$-means clustering and used the subsequent cluster assignments as supervisory information to update the weights of the neural network.
Huang et al. \cite{huang2020deep} proposed a deep clustering method termed PartItion Confidence mAximization (PICA),  which seeks to maximize the global partition confidence of the clustering solution. 
Besides the single-stage deep clustering methods \cite{yang2017towards,xie2016unsupervised,yang2016joint,guo2017improved,ghasedi2017deep,caron2018deep,ji2019invariant,huang2020deep}, another popular category is the two-stage deep clustering methods. Specifically, van Gansbeke et al. \cite{van2020scan} presented the SCAN method, which first conducts a pretext task of contrastive learning to mine the nearest neighbors, and then obtains the clustering result via the second-stage learning and clustering optimization. To extend the SCAN method, Dang et al. \cite{dang2021nearest} proposed the NNM method to match samples with their nearest neighbors from both local and global levels.

Further, some graph-based deep clustering methods \cite{peng2021attention,bo2020structural,chiang2019cluster} have recently been designed  to exploit the structural information underlying the data. For example, Chiang et al. \cite{chiang2019cluster} proposed a fast and memory-efficient deep clustering method based on Graph Convolutional Network (GCN). At each step, it samples a node block associated with a dense subgraph identified by the graph clustering algorithm and restricts the neighborhood search within that subgraph. Bo and Wang et al. \cite{bo2020structural} developed the structural deep clustering network (SDCN) to integrate the structural information into deep clustering by combining the GCN with the DEC framework. Peng et al. proposed \cite{peng2021attention} the Attention-driven Graph Clustering Network (AGCN) to dynamically aggregate the node attribute features and the topological graph features, and adaptively fuse the multi-scale features embedded at different layers.

Though significant achievements have been made, most of the previous deep clustering methods obtain the final clustering based on the single-layer representation in the neural network, which undermine their ability to effectively and comprehensively exploit the diverse information of data hidden in multi-layer representations.

\section{Proposed Approach}
\label{sec:proposed_approach}

In this section, our DeepCluE approach will be described in detail, which aims to harness the power of multiple layers in deep neural networks and jointly leverage the instance-level contrastive learning, the cluster-level contrastive learning, and the ensemble clustering via multi-layer collaboration in a unified deep clustering framework.

\subsection{Overview}
\label{sec:overview}

The pipeline of our DeepCluE approach is depicted in Fig.~\ref{fig:architecture}. Specifically, our DeepCluE approach mainly consists of two stages, i.e., the unsupervised contrastive network learning and the multi-layer ensemble clustering. The unsupervised training process of the contrastive network is implemented in an end-to-end manner with three main modules, including the weight sharing backbone network, the instance projector, and the cluster projector. Particularly, the backbone first extracts features from the sample pairs constructed through data augmentations on images. Then the instance projector and the cluster projector respectively perform contrastive learning in the row and column spaces of the feature matrix learned by the backbone. With the contrastive network trained, the base clusterings can be built on the feature representations from different layers in  different modules. Typically, multiple layers of feature representations from the instance projector, the cluster projector, and the backbone are jointly used. As the dimensions of different layers can be very different, we utilize the PCA to reduce the dimensions of some layers if their dimensions are greater than a threshold, e.g., 1000. Then, upon the feature representations extracted from multiple layers, we can generate a set of diversified base clusterings via the formulating and partitioning of multiple bipartite graphs built on multi-layer representations. To take into account the potentially different reliability of different base clusterings, an entropy-based criterion is utilized to estimate the local uncertainty of different ensemble members, based on which a unified weighted-cluster bipartite graph is constructed and then efficiently partitioned to achieve the final consensus clustering (as shown in Fig.~\ref{fig:architecture}).

\subsection{Unsupervised Contrastive Network Training}
\label{sec:CC}

In DeepCluE, we first utilize contrastive learning at both the instance-level and the cluster-level to train the deep neural network in an unsupervised manner \cite{li2021contrastive}, where the multiple layers of trained representations are then fed to the ensemble clustering process to build the final clustering. In this section, we describe the three modules in the contrastive network, namely, the backbone, the instance projector, and the cluster projector, in Sections~\ref{sec:backbone}, \ref{sec:instance_head}, and \ref{sec:cluster_head}, respectively.

\subsubsection{Backbone}
\label{sec:backbone}
To enforce the contrastive learning \cite{chen2020simple,chen2020improved,caron2020unsupervised,grill2020bootstrap}, different types of data augmentations are first performed on the images to generate the sample pairs. Specifically, given an image $x_i$, two data transformations $T^a$, $T^b$ randomly sampled from the same family of augmentations $\mathcal{T}$ are applied to this image, leading to two correlated views of $x_i$, which are denoted as $x^a_i = T^a(x_i)$ and $x^b_i = T^b(x_i)$, respectively. As suggested in \cite{chen2020simple}, the composition of multiple data augmentation operations is crucial to the representation learning performance in contrastive learning. In this work, we adopt an augmentation family with five types of data augmentation operations, namely, resized-crop, horizontal-flip, grayscale, color-jitter, and Gaussian-blur. Note that each augmentation is applied independently with a certain probability.
Thereafter, we utilize the ResNet34 \cite{he2016deep} as the weight-sharing backbone network to extract features from the two augmented samples, denoted as $h^a_i = f(x^a_i)$ and $h^b_i = f(x^b_i)$, respectively,  which are then fed to the instance projector and the cluster projector for the later contrastive learning.

\subsubsection{Instance-Level Contrastiveness}
\label{sec:instance_head}
How to define the positive and negative samples has always been a key problem in contrastive learning, which aims to maximize the similarities of the positive pairs and minimize that of the negative pairs by means of some contrastive loss. Typically, we randomly sample a mini-batch of $n$ samples and use the backbone network to extract features of the augmented pairs, which lead to a total of $2\cdot n$ augmented samples. Instead of sampling the negative samples explicitly, for a specific positive pair $\{x^a_i,x^b_i\}$, we treat the other $2\cdot(n - 1)$ augmented samples within a mini-batch as the negative pairs.

Specifically, following the backbone network, the instance projector $g_I(\cdot)$, a nonlinear multi-layer perceptron (MLP) with two fully-connected layers, is exploited to map the representations $h^a_i$ and $h^b_i$ to a low-dimensional subspace, denoted as $p^a_i = g_I(h^a_i)$ and $p^b_i = g_I(h^b_i)$, where the contrastive loss \cite{chen2020simple,li2021contrastive} is utilized for the instance-level contrastive learning. Previous studies \cite{chen2020simple} have suggested that it is beneficial to define the contrastive loss on $p^a_i$ and $p^b_i$ rather than $h^a_i$ and $h^b_i$, which can alleviate the information loss induced by the contrastive loss \cite{chen2020simple}.

The pairwise similarity is measured by the cosine similarity, denoted as $s(u,v) = (u^\top v)/(\|u\|\|v\|)$, where $u$ and $v$ are two feature vectors with the same dimension. For a given sample $x_i$, we optimize the pairwise similarity via the contrastive loss computed across all positive pairs, with both $\{x_i^a, x_i^b\}$ and $\{x_i^b, x_i^a\}$ taken into account, that is
\begin{equation}\label{eq1}
	l_i^a=-log\frac{\exp(s(p_i^a,p_i^b)/\tau_I)}{\sum_{j=1}^{n}[\exp(s(p_i^a,p_j^a)/\tau_I)+\exp(s(p_i^a,p_j^b)/\tau_I)]},
\end{equation}
\begin{equation}\label{eq2}
	l_i^b=-log\frac{\exp(s(p_i^b,p_i^a)/\tau_I)}{\sum_{j=1}^{n}[\exp(s(p_i^b,p_j^b)/\tau_I)+\exp(s(p_i^b,p_j^a)/\tau_I)]},
\end{equation}
where $\tau_I$ denotes the instance-level temperature parameter, and $l_i^a$ and $l_i^b$ denote the loss of sample $x_i$ w.r.t. the two random augmentations $T^a$ and $T^b$, respectively. Finally, the instance-level contrastive loss can be computed by traversing all augmented samples, that is

\begin{equation}\label{eq3}
	L_{con}=\frac{\sum_{i=1}^{n}(l_i^a+l_i^b)}{2n}.
\end{equation}

By means of the contrastive loss $L_{con}$, the optimization of the  instance-level contrastiveness can be achieved by pulling the similar instances (i.e., positive samples) closer and pushing dissimilar instances (i.e., negative samples) away.

\subsubsection{Cluster-Level Contrastiveness}
\label{sec:cluster_head}
By optimizing the instance-level contrastive loss, the similarity between individual samples is captured, while the cluster-level structure information is still unconsidered. Therefore, we further incorporate a cluster projector into the network. Note that the \emph{Softmax} fully-connected layer can realize the mapping of a sample to the cluster space, leading to the soft label whose the $i$-th element can be regarded as its probability of belonging to the $i$-th cluster. With the soft labels stacked to form a feature matrix, the global cluster information can be revealed \cite{li2021contrastive}.

Formally, let $D^a\in\mathbb{R}^{n\times K}$ be the output of cluster projector for the first augmentation (and $D^b\in\mathbb{R}^{n\times K}$ for the second augmentation), which can be obtained by stacking the soft labels of the $n$ samples, where $n$ and $K$ respectively denote the mini-batch size and the number of clusters.
Then we pay attention to the columns of the matrix $D^t$ (for $t\in\{a,b\}$), where the $i$-th column can be regarded as the vectorized representation of the $i$-th cluster. In this case, we expect that the two representations of the same cluster built through two different augmentations, respectively, should be close to each other, whereas the different clusters should be as dissimilar as possible, so as to maintain the consistency of the global cluster structure.

Specifically, following the backbone network, we further incorporate a cluster projector $g_C(\cdot)$, where a two-layer MLP with \emph{Softmax} is used to project the representations $h^a_i$ and $h^b_i$ into a $K$-dimensional space, leading to the soft assignments $\tilde{q}^a_i = g_C(h^a_i)$ and $\tilde{q}^b_i = g_C(h^b_i)$ for samples $x_i^a$ and $x_i^b$, respectively, where $\tilde{q}^t_i$ corresponds to the $i$-th \emph{row}  of $D^t$ for $t\in\{a,b\}$. Let $q_i^a$ denote the $i$-th \emph{column} of $D^a$,  corresponding to the representation of cluster $i$ under the first data augmentation. We match it with $q_i^b$ to form a positive cluster pair $\{q_i^a, q_i^b\}$, and use the other $2\cdot(K - 1)$ pairs as the negative pairs. For the $i$-th cluster, with both $\{q_i^a, q_i^b\}$ and $\{q_i^b, q_i^a\}$ considered, the cluster-level contrastive loss is defined as
\begin{equation}\label{eq4}
	\tilde{l}_i^a=-log\frac{\exp(s(q_i^a,q_i^b)/\tau_C)}{\sum_{j=1}^{K}[\exp(s(q_i^a,q_j^a)/\tau_C)+\exp(s(q_i^a,q_j^b)/\tau_C)]},
\end{equation}
\begin{equation}\label{eq5}
	\tilde{l}_i^b=-log\frac{\exp(s(q_i^b,q_i^a)/\tau_C)}{\sum_{j=1}^{K}[\exp(s(q_i^b,q_j^b)/\tau_C)+\exp(s(q_i^b,q_j^a)/\tau_C)]},
\end{equation}
where $\tau_C$ denotes the cluster temperature parameter. Thereby, the cluster-level contrastive loss w.r.t. the $K$ clusters can be computed as
\begin{equation}\label{eq6}
	L_{clu}=\frac{\sum_{i=1}^{K}(\tilde{l}_i^a+\tilde{l}_i^b)}{2K}-H(D),
\end{equation}
where the entropy term $H(D)$ is incorporated to prevent the trivial solution that assigns a majority of samples into one or a few clusters. This term also takes into account the representations under two augmentation operations, that is
\begin{equation}\label{eq7}
	H(D)=-{\textstyle \sum_{i=1}^{K}}[P(q_i^a)logP(q_i^a)+P(q_i^b)logP(q_i^b)],
\end{equation}
\begin{equation}\label{eq8}
	P(q_i^t)=\frac{{\textstyle \sum_{v=1}^{n}D_{vi}^t}}{\|D^t\|_1} ,~for~t\in\{a,b\}.
\end{equation}

By updating the network via the loss of $L_{clu}$, the cluster structure can be optimized via the cluster-level contrastiveness. Finally, we proceed to combine the optimization of the instance-level contrastiveness and the cluster-level contrastiveness into a unified loss function, that is
\begin{equation}\label{eq9}
	L_{total}=L_{con}+L_{clu}.
\end{equation}

With both the instance-level contrastiveness and the cluster-level contrastiveness leveraged, the overall network can be trained in a self-supervised (or unsupervised) manner. In this work, we aim to simultaneously take advantage of multiple layers of representations from multiple modules to enhance the clustering performance, which will be described in the following two sections.

\subsection{Diversified Ensemble Generation from Multiple Layers}
\label{sec:ensemble_generation}

The existing deep clustering studies \cite{yang2017towards,xie2016unsupervised,yang2016joint,guo2017improved,ghasedi2017deep,caron2018deep,ji2019invariant,Choudhury2021,huang2020deep,li2021contrastive,van2020scan,dang2021nearest} mostly build the final clustering by utilizing a single layer of the learned representation, e.g., by performing the $K$-means algorithm on the last fully-connected layer or by adding a \emph{Softmax} layer after the last fully-connected layer (which is then associated with some clustering loss). However, few of them have gone beyond the single-layer clustering paradigm to explore the more possibilities in multi-layer representations.

Inspired by the ensemble clustering technique \cite{fred2005combining,iam2011link,yi2012robust,huang2016ensemble,huang2017locally,huang2018enhanced,huang21_tcyb}, whose objective is to fuse multiple base clusterings for building a more robust clustering result, in this paper, we extends the conventional deep clustering framework from the single-layer clustering fashion to the multi-layer clustering fashion. Especially, we first focus on the problem of how to generate a set of diversified base clusterings from the multiple layers of representations in this section, and then deal with the fusion of these multiple base clusterings (via the weighted-cluster bipartite graph based consensus function) in the next section.

As illustrated in Fig.~\ref{fig:architecture}, we jointly utilize multiple layers of representations from three different modules, i.e., the backbone, the instance projector, and the cluster projector for our ensemble clustering process. In our framework, we adopt the ResNet34, a deep residual network with 34 weighted layers, as the backbone, which mainly consists of the following components, namely, the first convolutional layer $conv1$, the four residual structural modules $convi\_x$ ($i\in\{2,3,4,5\}$), and the average pooling layer $fc6$ with flatten operation. Besides the backbone, the instance projector $g_I(\cdot)$ consists of two fully-connected layers, i.e., $fci$ ($i\in\{7,8\}$), whereas the cluster projector $g_C(\cdot)$ also consists two fully-connected layers, i.e., $fci$ ($i\in\{9,10\}$), where the \emph{Softmax} operation is used in the last layer for producing the soft labels.

Note that \emph{multiple} layers of representations are extracted from each of the three modules for ensemble generation. Let $\lambda_B$, $\lambda_I$, and $\lambda_C$ denote the numbers of layers extracted from the backbone, the instance projector, and the cluster projector, respectively. Thus a total of $\lambda = \lambda_B+\lambda_I+\lambda_C$ layers of representations will be extracted from the entire network. Let $N$ be the number of samples in the dataset and $d^{(i)}$ be the dimension of the $i$-th extracted layer. If the dimension of a layer exceeds a certain threshold, e.g., 1000, then PCA will be utilized to reduce the dimension of the representation of this layer. Let $\mathcal{X} = \{x_1,\cdots,x_N\}$ be an image dataset with $N$ samples, where $x_i$ is the $i$-th image sample. Then, the feature matrix of the $j$-th extracted layer can be represented as $Y^{(j)}\in\mathbb{R}^{N\times d^{(j)}}$, where each row corresponds to the feature representation of a sample. For example, the $i$-th row in $Y^{(j)}$, denoted as $y^{(j)}_i\in\mathbb{R}^{d^{(j)}}$, corresponds to the feature representation of the sample $x_i$ in the $j$-th extracted layer. Thereby, we can represent the original dataset by the $\lambda$ extracted feature representations, that is
\begin{equation}
	\mathcal{Y} = \{Y^{(1)},\cdots,Y^{(\lambda)}\}.
\end{equation}

Instead of generating one base clustering at each layer, to inject diversity into the ensemble system, we produce \emph{multiple} diversified base clusterings at \emph{each} layer of feature representation. With consideration to both clustering quality and efficiency, we adopt the Ultra-scalable SPEctral Clustering (U-SPEC) algorithm \cite{huang19_tkde} for ensemble generation, which takes advantage of the bipartite graph structure and is featured by its linear time and space complexity in the sample size $N$. Specifically, it first selects a set of representatives via the hybrid representative selection strategy, then builds a bipartite graph between the original samples and the representatives via the fast approximation of $k$-nearest neighbors, and finally obtains a base clustering by efficiently partitioning the bipartite graph. By performing the U-SPEC algorithm multiple times, multiple base clusterings can be obtained from each layer of representation. Then, a question may arise as to how to diversify the base clusterings of multiple runs of U-SPEC, which in fact is addressed in two aspects. First, the hybrid representative selection requires random down-sampling and $K$-means clustering, which can lead to a different set of representatives at each run. Second, the number of clusters for each base clustering is randomly selected, which further enforces the diversity of the base clusterings.

Formally, let $M'$ denote the number of base clusterings generated at each extracted layer. Then a total of $M=\lambda\cdot M'$ base clusterings can be obtained.
The ensemble of base clusterings generated from multiple layers of representations can be denoted as
\begin{equation}\label{eq10}
	\Pi=\{\pi^1,\cdots,\pi^M\},
\end{equation}
where $\pi^m=\{C^m_1,\cdots,C^m_{k^m}\}$ is the $m$-th base clustering, $C^m_i$ is the $i$-th cluster in $\pi^m$, and $k^m$ is the number of clusters in $\pi^m$. Then the next question that remains to be tackled is how to effectively and efficiently fuse the information of the multiple base clusterings to build a probably more robust consensus clustering $\pi^*$.

\subsection{Weighted Bipartite Graph Based Consensus Function}
\label{sec:BGCF}

With the set of base clusterings generated by exploiting multiple layers in the deep neural network, in this section, we proceed to combine the base clusterings into the final clustering via the weighted-cluster bipartite graph based consensus function.

Diversity and quality are two crucial factors for ensemble clustering. Although different layers of representations can provide rich and diverse information for the clustering, yet the reliability of different layers (or even different base clusterings generated in the same layer) may be quite different. Thereby, before fusing multiple base clusterings (via ensemble clustering), we first estimate the quality of the base clusterings and design the weighting scheme accordingly. Especially, rather than treating each base clustering as an individual, we estimate the reliability of the clusters in each base clustering by taking into account the distribution of the cluster labels in the entire ensemble via an entropy-based criterion.

For convenience, we represent the set of clusters in the $M$ base clusterings as follows:
\begin{equation}\label{eq11}
	\mathcal{C}=\{C_1,\cdots,C_{k_c}\},
\end{equation}
where $C_i$ denotes the $i$-th cluster, and $k_c$ denotes the total number of clusters in $\Pi$. It is obvious that $k_c=\sum_{i=1}^{M}k^i$.
Each cluster consists of a set of data samples. Without supervision, if the data samples in a cluster within a base clustering are frequently grouped into the same cluster in the other base clusterings, which means that multiple base clusterings \emph{agree} that the samples in this cluster should be together, then this cluster can be regarded as more reliable. To measure the agreement (or disagreement) among multiple base clusterings, we take advantage of the concept of entropy \cite{huang2017locally,huang21_tcyb}, which provides a simple yet effective measure of uncertainty for the clusters. Given a cluster $C_i$ and a base clustering $\pi^m$, the uncertainty (or entropy) of the cluster $C_i$ w.r.t. the base clustering $\pi^m$ can be measured by considering how the samples in $C_i$ are partitioned in $\pi^m$, that is
\begin{align}
	H^m(C_i)&=-\sum_{j=1}^{k^m}P(C_i,C^m_j)\log_2P(C_i,C^m_j),\\
	&P(C_i,C^m_j) = \frac{|C_i\bigcap C^m_j|}{|C_i|},
\end{align}
where $\bigcap$ denotes the intersection of two sets and $|\cdot|$ obtains the number of samples in a set.
Based on the assumption that the multiple base clusterings are independent of each other, the uncertainty of the cluster $C_i$ w.r.t. the entire ensemble $\Pi$ with $M$ base clusterings can be computed as follows:
\begin{align}
	H^*(C_i) = \sum_{m=1}^M H^m(C_i).
\end{align}

When the samples in $C_i$ belong to the same cluster in all the $M$ base clusterings, the uncertainty of $C_i$ w.r.t. the ensemble $\Pi$ reaches its minimum value zero. 
Thus, with the uncertainty of clusters defined, we can further present the cluster-wise weighting scheme for our bipartite graph based consensus function.

By treating both data samples and base clusters as nodes, we can define the weighted-cluster bipartite graph for the clustering ensemble $\Pi$ as follows:
\begin{equation}\label{eq12}
	G=\left \{\mathcal{X},C,B\right \},
\end{equation}
where $\mathcal{X}\bigcup C$ represents the node set and $B$ represents the cross-affinity matrix.  The reason for constructing a bipartite graph rather than a general graph is two-fold. First, for the ensemble of multiple base clusterings, the bipartite graph can naturally encode the relationship between the original samples and the base clusters. Second, in comparison with a general graph with an $N\times N$ similarity matrix, the bipartite graph can be partitioned in a more efficient manner. In the following, we will further incorporate the uncertain (or reliablity) of clusters into the bipartite graph structure.

As the uncertainty of different clusters has been estimated via the entropy-based criterion, it is expected that a cluster with higher reliability (corresponding to lower uncertainty) should exert a greater influence. Therefore, we can define the weight of a cluster $C_i$ by considering its uncertainty, that is
\begin{align}
	w(C_i) = \exp\left(-\frac{H^*(C_i)}{M}\right), ~\text{for}~ C_i\in \mathcal{C}.
\end{align}

It holds that $w(C_i)\in (0,1]$. When the uncertainty of $C_i$ reaches its minimum value zero, its weight $w(C_i)$ reaches its maximum value one.
Thus, the cross-affinity matrix for the weighted-cluster bipartite graph can be defined as follows:
\begin{equation}\label{eq13}
	B=\{b_{ij}\}_{N\times k_c},
\end{equation}
\begin{equation}\label{eq14}
	b_{ij}=\begin{cases}
		w(C_j),&\text{if}\ x_i\in C_j \\
		0,&\text{otherwise.}
	\end{cases}
\end{equation}

\begin{algorithm}[!t]
	\label{alg}
	\caption{\mbox{Deep Clustering via Ensembles (DeepCluE)}}
	\LinesNumbered
	\KwIn{Dataset $\mathcal{X}$; Backbone $f(\cdot)$; Instance projector $g_I(\cdot)$; Cluster projector $g_C(\cdot)$; Temperature parameters $\tau_I$ and $\tau_C$; Augmentation group $\mathcal{T}$; Training epochs $E$; Batch size $n$; Number of clusters $K$; Number of extracted layers $\lambda$; Number of base clusterings per layer $M'$.} 
	\KwOut{Final clustering result $\pi^*$.}
	\# Training contrastive network according to Sec.\ref{sec:CC}\;
	\For{epoch = $1$,...,$E$}{
		Sample a mini-batch $\{x_i\}_{i=1}^n$ from $\mathcal{X}$\;
		Sample two augmentations $T^a,T^b$ from $\mathcal{T}$\;
		Feed $T^a(x_i),T^b(x_i)$ into the model\;
		Compute $\mathcal{L}_{con},\mathcal{L}_{clu}$ and $\mathcal{L}_{total}$\;
		Update the parameters of $f(\cdot)$, $g_I(\cdot)$, and $g_C(\cdot)$\;
	}
	\# Ensemble generation according to Sec.\ref{sec:ensemble_generation}\;
	Initialize an empty ensemble $\Pi$\;
	\For{all $Y^{(i)}\in \mathcal{Y}$}{
		\# For each layer, generate $M'$ base clusterings\;
		\For{$m$ = $1$,...,$M'$}{
			Initialize a random cluster number $k^m$\;
			Obtain a base clustering $\pi^m$ via U-SPEC\;
			Update the ensemble: $\Pi=\Pi\cup\pi^m$\;
		}
	}
	\# Weighted consensus function according to Sec.\ref{sec:BGCF}\;
	Estimate the uncertainty of each cluster\;
	Compute the weight for each cluster\;
	Build the weighted-cluster bipartite graph $G$ for $\Pi$\;
	Partition $G$ into $K$ disjoint subsets via Tcut\;
	\For{all $x_i\in \mathcal{X}$}{
		Assign a cluster label to $x_i$ according to which subset it is in\;
	}
\end{algorithm}

In the constructed bipartite graph, an edge between two graph nodes exists if and only if one of them is a data sample and the other is the cluster that contains it, whose weight is decided by the reliability of the corresponding cluster. Due to the imbalanced structure of the bipartite graph, with $N\gg k_c$, the transfer cut (Tcut) \cite{li2012segmentation} can be adopted to partition the graph efficiently and thus obtain the final clustering, whose computational complexity is linear in $N$ and cubic in $k_c$.

For clarity, the overall process of our DeepCluE algorithm is summarized in Algorithm~\ref{alg}.

\section{Experiments}
\label{sec:experiments}

In this section, we experimentally evaluate the clustering performance of our DeepCluE approach against the state-of-the-art deep clustering approaches on multiple image datasets.

\subsection{Datasets and Evaluation Metrics}

\begin{figure}[!t]
	\begin{center}
		{\subfigure[Fashion]
			{\includegraphics[width=0.95\columnwidth]{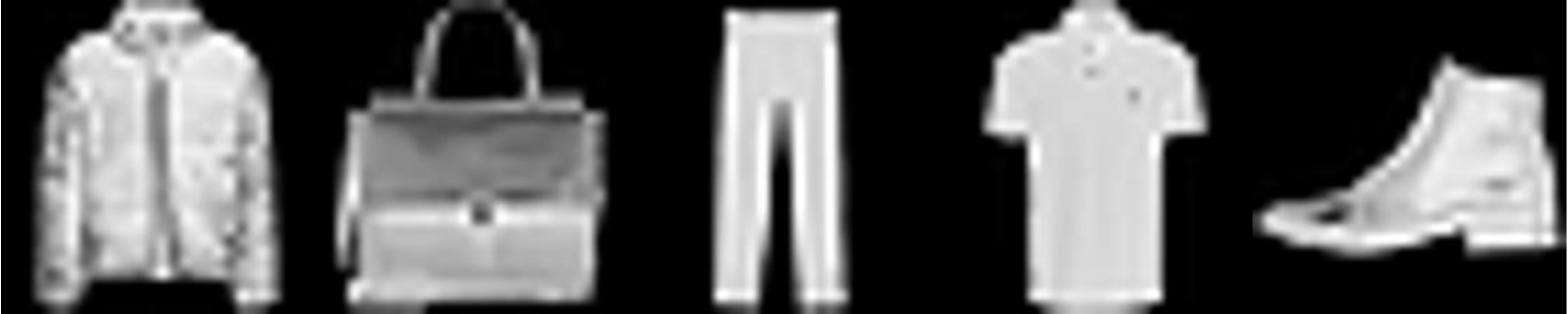}}}\\
		{\subfigure[CIFAR-10]
			{\includegraphics[width=0.95\columnwidth]{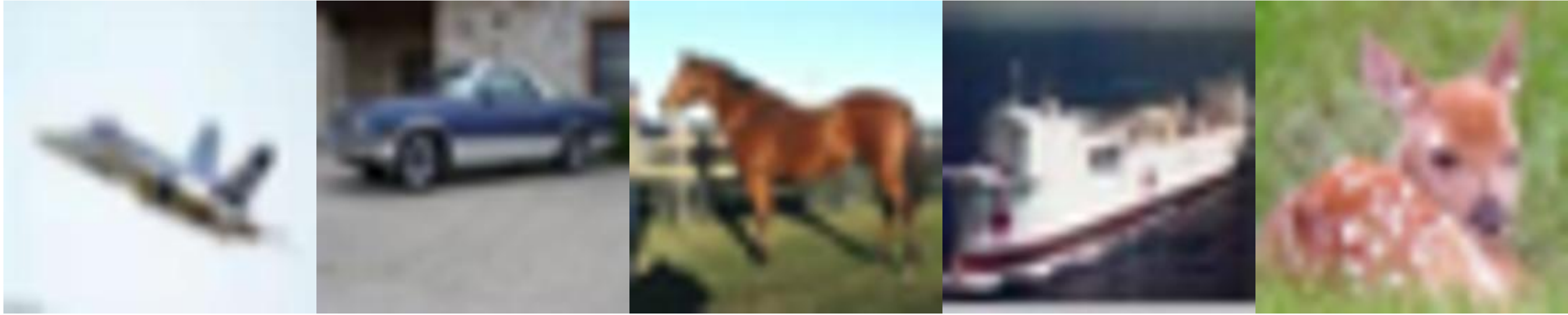}}}\\
		{\subfigure[CIFAR-100]
			{\includegraphics[width=0.95\columnwidth]{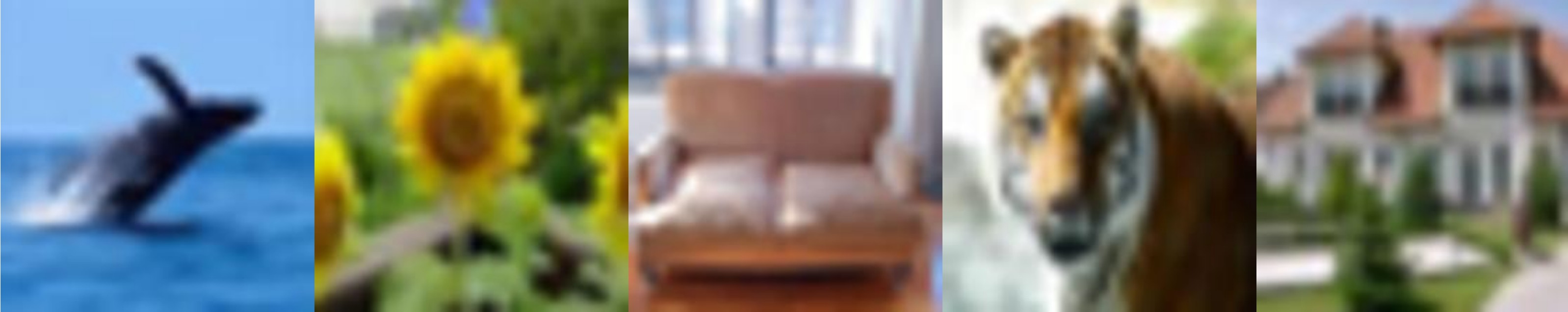}}}\\
		{\subfigure[ImageNet-10]
			{\includegraphics[width=0.95\columnwidth]{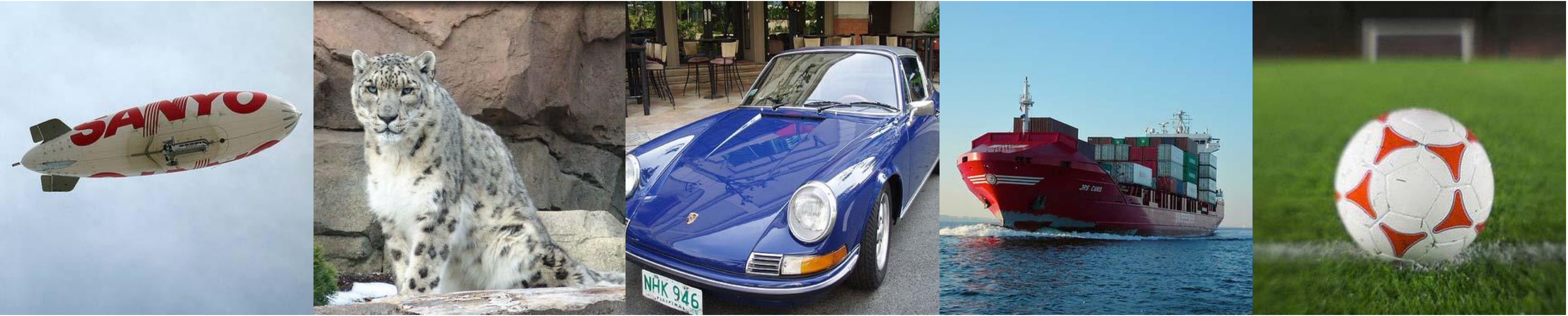}}}\\
		{\subfigure[ImageNet-Dogs]
			{\includegraphics[width=0.95\columnwidth]{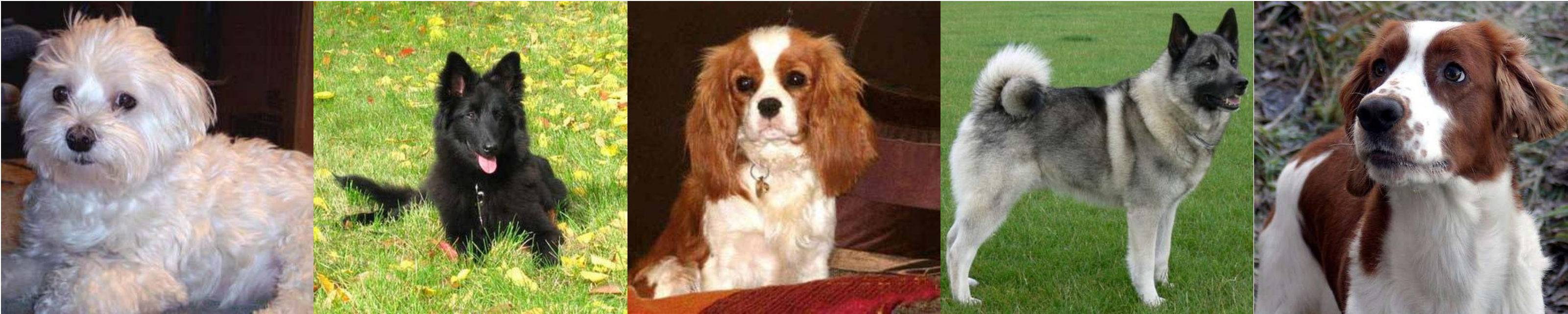}}}\\
		{\subfigure[Tiny-ImageNet]
			{\includegraphics[width=0.95\columnwidth]{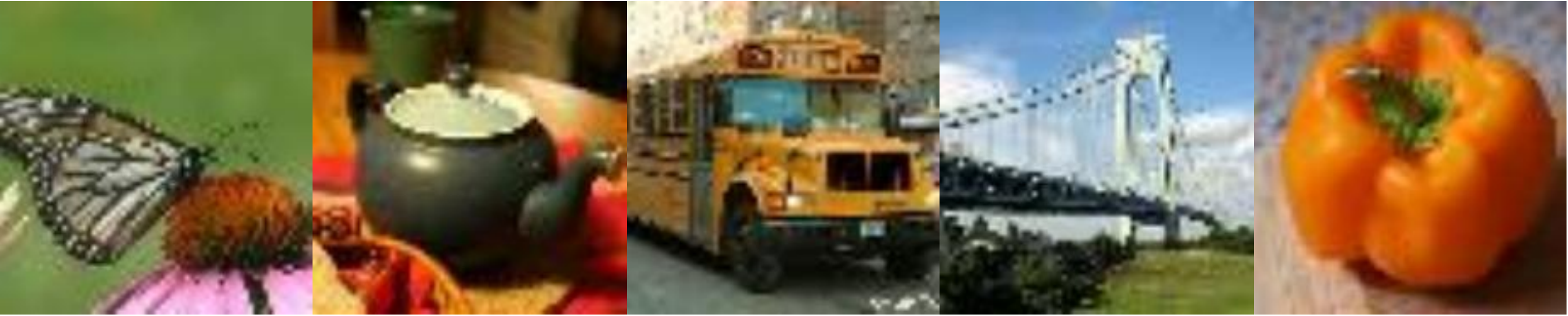}}}\\
		\caption{Some examples of the six image datasets. Note that the Fashion dataset is a gray-scale image dataset,  while the other five datasets are three-channel color image datasets.}
		\label{fig:datasets}
	\end{center}
\end{figure}

We conduct experiments on six image datasets for image clustering, namely, Fashion \cite{xiao2017fashion}, CIFAR-10 \cite{krizhevsky2009learning}, CIFAR-100 \cite{krizhevsky2009learning}, ImageNet-10 \cite{li2021contrastive}, ImageNet-Dogs \cite{li2021contrastive}, and Tiny-ImageNet \cite{deng2009imagenet}.
The statistics of these benchmark datasets are provided in Table~\ref{table:dataset}, and some examples in these datasets are visualized in Fig. \ref{fig:datasets}.

\begin{table}[!t]
	\small
	\centering
	\caption{Statistics of the Datasets.}
	\label{table:dataset}
	\begin{tabular}{m{3.3cm}<{\centering}m{1.8cm}<{\centering}m{1.7cm}<{\centering}}
		\toprule[1pt]
		Dataset &\#Samples &\#Classes\\
		\midrule
		Fashion  &70,000   &10\\
		CIFAR-10  &60,000   &10\\
		CIFAR-100  &60,000   &20\\
		ImageNet-10  &13,000   &10\\
		ImageNet-Dogs  &19,500   &15\\
		Tiny-ImageNet  &100,000   &200\\
		\bottomrule[1pt]
	\end{tabular}
\end{table}

To quantitatively evaluate the clustering quality, three widely-adopted metrics are adopted, namely, normalized mutual information (NMI) \cite{yang22_tetci}, adjusted Rand index (ARI) \cite{Liang2022}, and clustering accuracy (ACC) \cite{Fang2023}.

\begin{table*}[!t]
	\caption{The NMI scores of different clustering methods on six image datasets. The best score on each dataset is in \textbf{Bold}.}
	\label{table:nmi_results}
	\small
	\centering
	\begin{tabular}{|m{2.1cm}<{\centering}|m{2cm}<{\centering}|m{2cm}<{\centering}|m{2cm}<{\centering}
			|m{2cm}<{\centering}|m{2.1cm}<{\centering}|m{2cm}<{\centering}|}
		\hline
		Dataset & Fashion & CIFAR-10 & CIFAR-100 & ImageNet-10 & ImageNet-Dogs & Tiny-ImageNet\\
		\hline\hline
		$K$-means \cite{jain2010data}	           &0.512&0.087 &0.084  &0.119 &0.055 &0.065\\\hline
		SC \cite{ng2002spectral}	           &0.659&0.103 &0.090	&0.151 &0.038 &0.063\\\hline
		AC \cite{jain2010data}	               &0.564&0.105 &0.098	&0.138 &0.037 &0.069\\\hline
		NMF	\cite{cai2009locality}             &0.425&0.081 &0.079	&0.132 &0.044 &0.072\\\hline
		AE \cite{bengio2006greedy}	           &0.567&0.239 &0.100	&0.210 &0.104 &0.131\\\hline
		DAE	\cite{vincent2010stacked}          &-	 &0.251 &0.111	&0.206 &0.104 &0.127\\\hline
		DCGAN \cite{radford2015unsupervised}   &-	 &0.265 &0.120	&0.225 &0.121 &0.135\\\hline
		DeCNN \cite{zeiler2010deconvolutional} &-	 &0.240 &0.092	&0.186 &0.098 &0.111\\\hline
		VAE	\cite{kingma2013auto}              &-	 &0.245 &0.108	&0.193 &0.107 &0.113\\\hline
		JULE \cite{yang2016joint}	           &0.608&0.192	&0.103	&0.175 &0.054 &0.102\\\hline
		DEC \cite{xie2016unsupervised}	       &0.601&0.257	&0.136	&0.282 &0.122 &0.115\\\hline
		DAC \cite{chang2017deep}	           &0.632&0.396	&0.185	&0.394 &0.219 &0.190\\\hline
		DCCM \cite{wu2019deep}	               &-	 &0.496	&0.285	&0.608 &0.321 &0.224\\\hline
		IIC \cite{ji2019invariant}	           &0.637	 &0.513	&-	&- &- &-\\\hline
		PICA \cite{huang2020deep}	           &-	 &0.591	&0.310	&0.802 &0.352 &0.277\\\hline
		CC \cite{li2021contrastive}	           &0.641&0.705	&0.430	&0.862 &0.401 &0.314\\\hline
		CLD \cite{wang2021unsupervised}	           &0.532&0.443&0.425&0.671&0.279 &0.308\\\hline
		HCSC \cite{guo2022hcsc}	           &0.472&0.407&0.361&0.647&0.355 &0.305\\\hline\hline
		\textbf{DeepCluE} &\textbf{0.694} &\textbf{0.727} &\textbf{0.472} &\textbf{0.882} &\textbf{0.448}	&\textbf{0.379}\\
		\hline
	\end{tabular}
\end{table*}

\begin{table*}[!t]
	\caption{The ARI scores of different clustering methods on six image datasets. The best score on each dataset is in \textbf{Bold}.}
	\label{table:ari_results}
	\small
	\centering
	\begin{tabular}{|m{2.1cm}<{\centering}|m{2cm}<{\centering}|m{2cm}<{\centering}|m{2cm}<{\centering}
			|m{2cm}<{\centering}|m{2.1cm}<{\centering}|m{2cm}<{\centering}|}
		\hline
		Dataset & Fashion & CIFAR-10 & CIFAR-100 & ImageNet-10 & ImageNet-Dogs & Tiny-ImageNet\\
		\hline\hline
		$K$-means \cite{jain2010data}	           &0.348&0.049	&0.028	&0.057	&0.020	&0.005\\\hline
		SC \cite{ng2002spectral}	           &0.468	 &0.085	&0.022	&0.076	&0.013	&0.004\\\hline
		AC \cite{jain2010data} 	               &0.371&0.065	&0.034	&0.067	&0.021	&0.005\\\hline
		NMF \cite{cai2009locality}	           &-	 &0.034	&0.026	&0.065	&0.016	&0.005\\\hline
		AE \cite{bengio2006greedy}	           &0.391&0.169	&0.048	&0.152	&0.073	&0.007\\\hline
		DAE	\cite{vincent2010stacked}          &-	 &0.163	&0.046	&0.138	&0.078	&0.007\\\hline
		DCGAN \cite{radford2015unsupervised}   &-	 &0.176	&0.045	&0.157	&0.078	&0.007\\\hline
		DeCNN \cite{zeiler2010deconvolutional} &-	 &0.174	&0.038	&0.142	&0.073	&0.006\\\hline
		VAE \cite{kingma2013auto}	           &-	 &0.167	&0.040	&0.168	&0.079	&0.006\\\hline
		JULE \cite{yang2016joint}	           &-	 &0.138	&0.033	&0.138	&0.028	&0.006\\\hline
		DEC \cite{xie2016unsupervised}	       &0.446&0.161	&0.050	&0.203	&0.079	&0.007\\\hline
		DAC	\cite{chang2017deep}               &0.502&0.306	&0.088	&0.302	&0.111	&0.017\\\hline
		DCCM \cite{wu2019deep}	               &-	 &0.408	&0.173	&0.555	&0.182	&0.038\\\hline
		IIC \cite{ji2019invariant}	           &0.523	 &0.411	&-	&- &- &-\\\hline
		PICA \cite{huang2020deep}	           &-	 &0.512	&0.171	&0.761	&0.201	&0.040\\\hline
		CC \cite{li2021contrastive}	           &0.545 &0.637	&0.266	&0.825	&0.225	&0.073\\\hline
		CLD \cite{wang2021unsupervised}	           &0.315&0.319&0.264&0.626&0.141 &0.061\\\hline
		HCSC \cite{guo2022hcsc}	           &0.279&0.295&0.206&0.559&0.209 &0.060\\\hline\hline
		\textbf{DeepCluE} &\textbf{0.569} &\textbf{0.646} &\textbf{0.288} &\textbf{0.856} &\textbf{0.273}	&\textbf{0.102}\\
		\hline
	\end{tabular}
\end{table*}

\begin{table*}[!t]
	\caption{The ACC scores of different clustering methods on six image datasets. The best score on each dataset is in \textbf{Bold}.}
	\label{table:acc_results}
	\small
	\centering
	\begin{tabular}{|m{2.1cm}<{\centering}|m{2cm}<{\centering}|m{2cm}<{\centering}|m{2cm}<{\centering}
			|m{2cm}<{\centering}|m{2.1cm}<{\centering}|m{2cm}<{\centering}|}
		\hline
		Dataset & Fashion & CIFAR-10 & CIFAR-100 & ImageNet-10 & ImageNet-Dogs & Tiny-ImageNet\\
		\hline\hline
		$K$-means \cite{jain2010data}	           &0.474 &0.229&0.130	&0.241	&0.105	&0.025\\\hline
		SC \cite{ng2002spectral}	           &0.583 &0.247&0.136	&0.274	&0.111	&0.022\\\hline
		AC \cite{jain2010data}                 &0.500 &0.228&0.138	&0.242	&0.139	&0.027\\\hline
		NMF \cite{cai2009locality}	           &0.434 &0.190&0.118	&0.230	&0.118	&0.029\\\hline
		AE \cite{bengio2006greedy}	           &0.540 &0.314&0.165	&0.317	&0.185	&0.041\\\hline
		DAE \cite{vincent2010stacked}	       &-	  &0.297&0.151	&0.304	&0.190	&0.039\\\hline
		DCGAN \cite{radford2015unsupervised}   &-	  &0.315&0.151	&0.346	&0.174	&0.041\\\hline
		DeCNN \cite{zeiler2010deconvolutional} &-	  &0.282&0.133	&0.313	&0.175	&0.035\\\hline
		VAE \cite{kingma2013auto}	           &-	  &0.291&0.152	&0.334	&0.179	&0.036\\\hline
		JULE \cite{yang2016joint}	           &0.563 &0.272&0.137	&0.300	&0.138	&0.033\\\hline
		DEC \cite{xie2016unsupervised}	       &0.590 &0.301&0.185	&0.381	&0.195	&0.037\\\hline
		DAC \cite{chang2017deep}	           &0.615 &0.522&0.238	&0.527	&0.275	&0.066\\\hline
		DCCM \cite{wu2019deep}	               &-	  &0.623&0.327	&0.710	&0.383	&0.108\\\hline
		IIC \cite{ji2019invariant}	           &0.657	 &0.617	&0.257	&- &- &-\\\hline
		PICA \cite{huang2020deep}	           &-	  &0.696&0.337	&0.870	&0.352	&0.098\\\hline
		CC \cite{li2021contrastive}	           &0.656 &$\textbf{0.790}$ &0.429 &0.895 &0.342	&0.136\\\hline
		CLD \cite{wang2021unsupervised}	           &0.495&0.542&0.420&0.807&0.315 &0.141\\\hline
		HCSC \cite{guo2022hcsc}	           &0.454&0.480&0.362&0.741&0.355 &0.139\\\hline\hline
		\textbf{DeepCluE} &\textbf{0.689} &0.764 &\textbf{0.457} &\textbf{0.924} &\textbf{0.416}	&\textbf{0.194}\\
		\hline
	\end{tabular}
\end{table*}

\subsection{Experimental Settings}

Different from many deep clustering methods that need to be fine-tuned for different datasets, our DeepCluE method doesn't require dataset-specific hyper-parameter-tuning, and is able to obtain consistently high-quality clustering results by using the same experimental setting on various datasets.

Specifically, with the ResNet34 adopted as the backbone, all input images are resized to a size of $224\times224$.
An augmentation family with five types of data augmentation operations, namely, resized-crop, horizontal-flip, grayscale, color-jitter, and Gaussian-blur, is utilized.
We take the Gaussian-blur augmentation out for the low-resolution datasets, including Fashion, CIFAR-10, CIFAR-100, and Tiny-ImageNet, since the up-scaling already results in blurred images. The two-layer MLP instance projector $g_I(\cdot)$ maps the representation to a 128-dimensional latent space, whereas the dimension of the output vector from the cluster projector $g_C(\cdot)$ is set to the cluster number. The instance temperature parameter $\tau_I$ is fixed to 0.5 and the cluster temperature parameter $\tau_C$ is fixed to $1.0$ in all experiments. The batch size is set to 256. We use the Adam optimizer \cite{kingma2014adam} with a learning rate of $3\times10^{-4}$ without weight decay or scheduler. Our model is trained for 1000 epochs in an unsupervised manner. We then extract $\lambda_B=3$ layers from the backbone, $\lambda_I=2$ layers from the instance projector, and $\lambda_C=1$ layer from the cluster projector.
Since the second MLP layer in the cluster projector is associated with $Softmax$, we only extract its first MLP layer. Therefore, a total of $\lambda = 6$ layers are utilized for ensemble clustering.
If the dimension of an extracted layer is greater than 1000, it will be PCA-reduced to 1000-dimensional. For each layer, $M'=5$ base clusterings are generated with the cluster number randomly chosen in $[K,\sqrt{N}]$, where $K$ and $N$ are respectively the number of clusters and the size of the dataset.

\begin{table*}[!t]
	\caption{The NMI Performance of Using A Single Layer and Using Multiple Layers.}
	\label{table:nmi_single_vs_multiple}
	\small
	\centering
	\begin{tabular}{|m{3.3cm}<{\centering}|m{1.5cm}<{\centering}|m{1.5cm}<{\centering}|m{1.7cm}<{\centering}|m{1.7cm}<{\centering}|m{2.1cm}<{\centering}|m{2cm}<{\centering}|}
		\hline
		Dataset & Fashion & CIFAR-10 & CIFAR-100 & ImageNet-10 & ImageNet-Dogs & Tiny-ImageNet\\
		\hline
		Using last layer of $g_I(\cdot)$ &0.565	&0.631 &0.409 &0.469 &0.354	&0.378\\
		\hline
		Using last layer of $g_C(\cdot)$ &0.641	&0.705 &0.430 &0.862 &0.401	&0.314\\
		\hline
		Using multiple layers	&\textbf{0.694}	&\textbf{0.727}	&\textbf{0.472}	&\textbf{0.882}	&\textbf{0.448}	&\textbf{0.379}\\
		\hline
	\end{tabular}
\end{table*}

\begin{table*}[!t]
	\caption{The ARI Performance of Using A Single Layer and Using Multiple Layers.}
	\label{table:ari_single_vs_multiple}
	\small
	\centering
	\begin{tabular}{|m{3.3cm}<{\centering}|m{1.5cm}<{\centering}|m{1.5cm}<{\centering}|m{1.7cm}<{\centering}|m{1.7cm}<{\centering}|m{2.1cm}<{\centering}|m{2cm}<{\centering}|}
		\hline
		Dataset & Fashion & CIFAR-10 & CIFAR-100 & ImageNet-10 & ImageNet-Dogs & Tiny-ImageNet\\
		\hline
		Using last layer of $g_I(\cdot)$ &0.406	&0.490 &0.190 &0.177 &0.135	&0.092\\
		\hline
		Using last layer of $g_C(\cdot)$ &0.545	&0.637 &0.266 &0.825 &0.225	&0.073\\
		\hline
		Using multiple layers	&\textbf{0.569}	&\textbf{0.646}	&\textbf{0.288}	&\textbf{0.856}	&\textbf{0.273}	&\textbf{0.102}\\
		\hline
	\end{tabular}
\end{table*}

\begin{table*}[!t]
	\caption{The ACC Performance of Using A Single Layer and Using Multiple Layers.}
	\label{table:acc_single_vs_multiple}
	\small
	\centering
	\begin{tabular}{|m{3.3cm}<{\centering}|m{1.5cm}<{\centering}|m{1.5cm}<{\centering}|m{1.7cm}<{\centering}|m{1.7cm}<{\centering}|m{2.1cm}<{\centering}|m{2cm}<{\centering}|}
		\hline
		Dataset & Fashion & CIFAR-10 & CIFAR-100 & ImageNet-10 & ImageNet-Dogs & Tiny-ImageNet\\
		\hline
		Using last layer of $g_I(\cdot)$ &0.526	&0.638 &0.382 &0.450 &0.344	&0.189\\
		\hline
		Using last layer of $g_C(\cdot)$ &0.656	&\textbf{0.790}	&0.429 &0.895 &0.342 &0.136\\
		\hline
		Using multiple layers	&\textbf{0.689}	&0.764	&\textbf{0.457}	&\textbf{0.924}	&\textbf{0.416}	&\textbf{0.194}\\
		\hline
	\end{tabular}
\end{table*}

\subsection{Compared with Traditional and Deep Clustering Methods}

In this section, we compare DeepCluE with eighteen baseline clustering methods, which include four traditional clustering methods, namely, $K$-means \cite{jain2010data}, Spectral Clustering (SC) \cite{ng2002spectral}, agglomerative clustering (AC) \cite{jain2010data}, and Nonnegative Matrix Factorization (NMF) \cite{cai2009locality}, and fourteen deep clustering methods, namely,
Auto-Encoder (AE) \cite{bengio2006greedy}, Denoising Auto-Encoder (DAE) \cite{vincent2010stacked}, Deep Convolutional Generative Adversarial Networks (DCGAN) \cite{radford2015unsupervised}, DeConvolutional Neural Networks (DeCNN) \cite{zeiler2010deconvolutional}, Variational Auto-Encoder (VAE) \cite{kingma2013auto}, Jointly Unsupervised LEarning (JULE) \cite{yang2016joint}, DEC \cite{xie2016unsupervised}, Deep Adaptive image Clustering (DAC) \cite{chang2017deep}, Deep Comprehensive Correlation Mining (DCCM) \cite{wu2019deep}, IIC \cite{ji2019invariant}, PICA \cite{huang2020deep}, CC \cite{li2021contrastive}, cross-level discrimination (CLD) \cite{wang2021unsupervised} and Hierarchical Contrastive Selective Coding (HCSC)\cite{guo2022hcsc}. The results of CC, CLD and HCSC are reproduced by using the authors' code, while the results of other baselines are taken from the corresponding papers.

The clustering performance w.r.t. NMI, ARI, and ACC of different clustering methods are reported in Tables~\ref{table:nmi_results}, \ref{table:ari_results}, and \ref{table:acc_results}, respectively. Note that previous deep clustering methods generally rely on single-layer output for generating the final clustering result. In comparison with the previous deep clustering methods, we find that our DeepCluE method with multiple layers of information jointly exploited can lead to better or significantly better clusterings on most of the benchmark datasets. In terms of NMI, as can be observed in Table~\ref{table:nmi_results}, DeepCluE  yields the best score on all the six image datasets. Especially, on the one hand, the deep clustering methods have exhibited significant advantages over the traditional clustering methods on most of the datasets, due to the representation learning ability of deep neural networks. On the other hand, our DeepCluE method consistently outperforms the other deep clustering methods. Specifically, on the CIFAR-100, ImageNet-Dogs, and Tiny-ImageNet datasets, our DeepCluE method achieves NMI scores of 0.472, 0.448, and 0.379, respectively, which significantly outperforms the best baseline method which achieves NMI scores of 0.430, 0.401, and 0.314, respectively. In terms of ARI and ACC, similar advantages can also be observed, which demonstrate the highly-competitive clustering performance of our DeepCluE method.

\subsection{Comparison of Single Layer Vs Multiple Layers}
\label{sec:comparison_single_multiple}

To evaluate the influence of using multiple layers simultaneously, in this section, we compare our DeepCluE method (using multiple layers) against the other two variants of only using a single layer on the benchmark datasets. As there are two projectors in our network, namely, the instance projector $g_I(\cdot)$ and the cluster projector $g_C(\cdot)$, the variants of only using the last layer of $g_I(\cdot)$ and only using the last layer of $g_C(\cdot)$ are respectively tested. When a single layer is used, we exploit the $K$-means clustering on this layer to produce the clustering. As shown in Tables~\ref{table:nmi_single_vs_multiple}, \ref{table:ari_single_vs_multiple}, and \ref{table:acc_single_vs_multiple}, the clustering performance of our DeepCluE method using multiple layers is much better than that of only using a single layer $g_I(\cdot)$ or  $g_C(\cdot)$, which shows the improvement brought in by jointly using multi-layer feature representations.

\begin{figure}[!t]
	\begin{center}
		{\subfigure[Fashion]
			{\includegraphics[width=0.26\columnwidth]{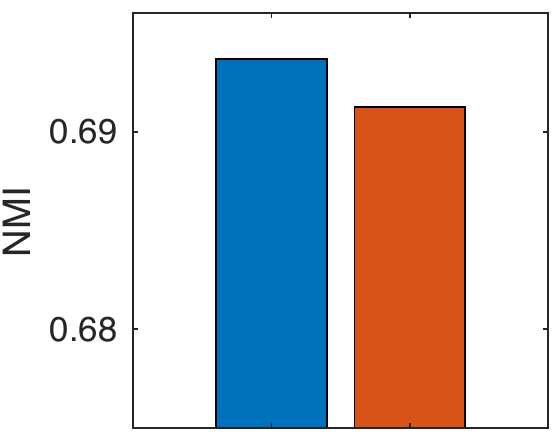}}}
		{\subfigure[CIFAR-10]
			{\includegraphics[width=0.26\columnwidth]{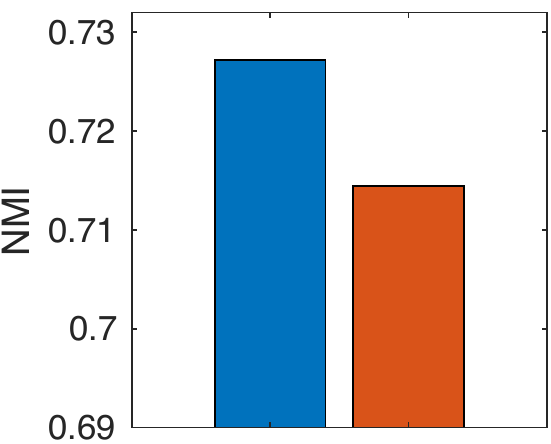}}}
		{\subfigure[CIFAR-100]
			{\includegraphics[width=0.26\columnwidth]{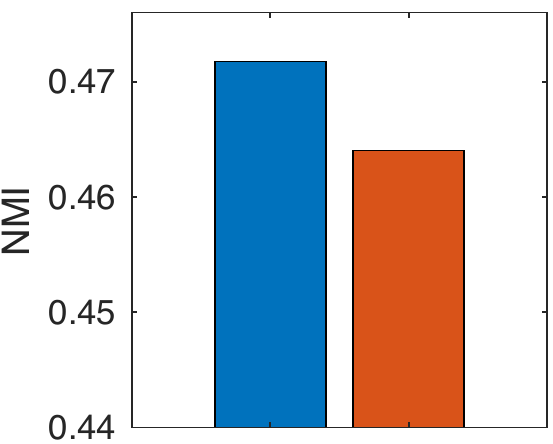}}}\\
		{\subfigure[ImageNet-10]
			{\includegraphics[width=0.26\columnwidth]{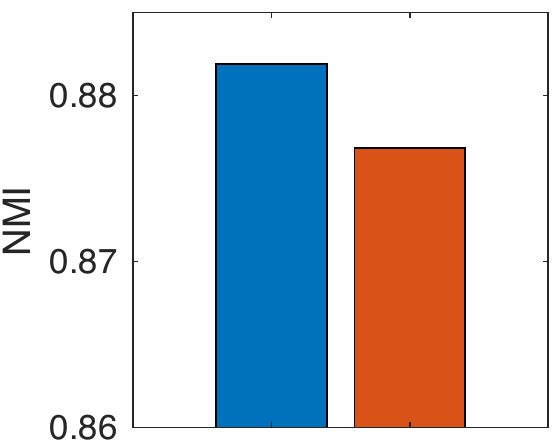}}}
		{\subfigure[ImageNet-Dogs]
			{\includegraphics[width=0.26\columnwidth]{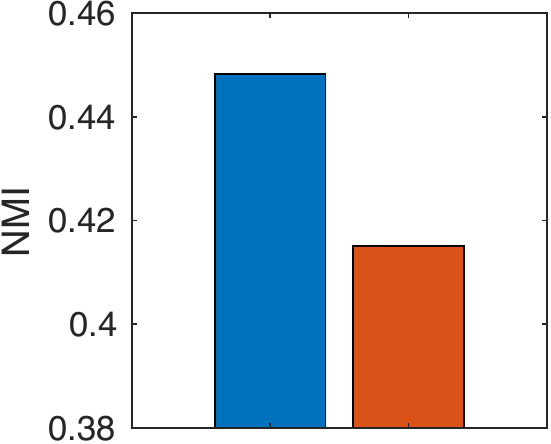}}}
		{\subfigure[Tiny-ImageNet]
			{\includegraphics[width=0.26\columnwidth]{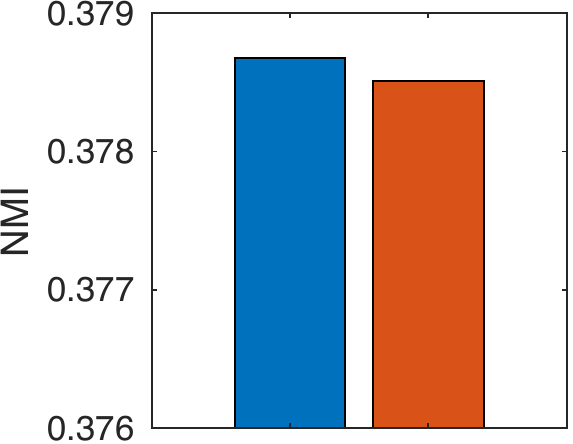}}}
		{\subfigure
			{\includegraphics[width=0.65\columnwidth]{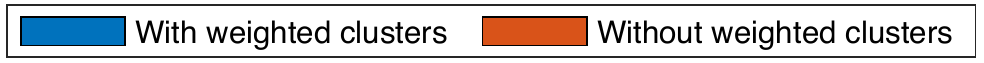}}}
		\caption{The NMI performance of DeepCluE with or without weighted clusters.}
		\label{fig:weighted_clusters_nmi}
	\end{center}
\end{figure}

\begin{figure}[!t]
	\begin{center}
		{\subfigure[Fashion]
			{\includegraphics[width=0.26\columnwidth]{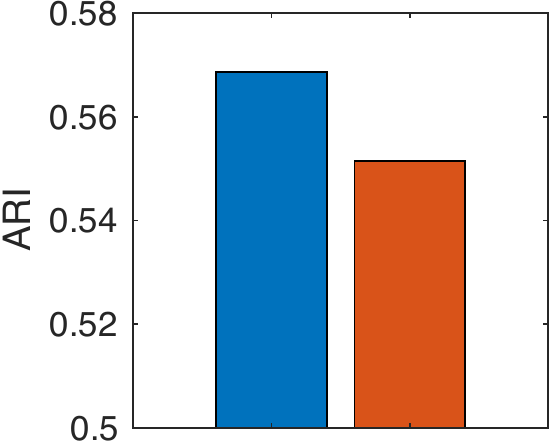}}}
		{\subfigure[CIFAR-10]
			{\includegraphics[width=0.26\columnwidth]{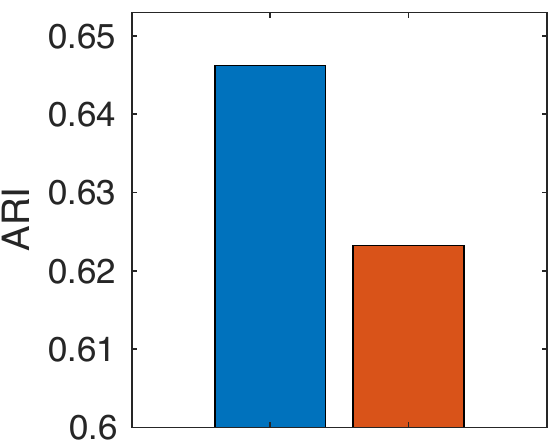}}}
		{\subfigure[CIFAR-100]
			{\includegraphics[width=0.26\columnwidth]{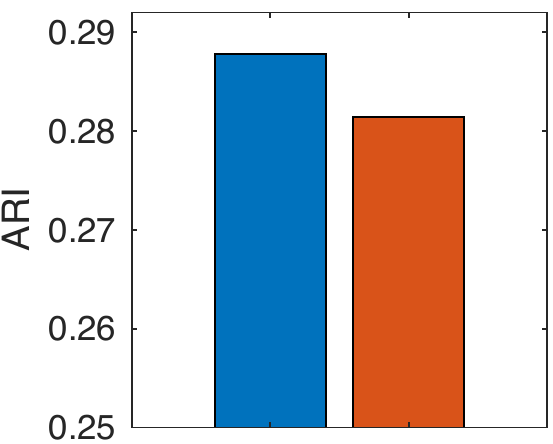}}}\\
		{\subfigure[ImageNet-10]
			{\includegraphics[width=0.26\columnwidth]{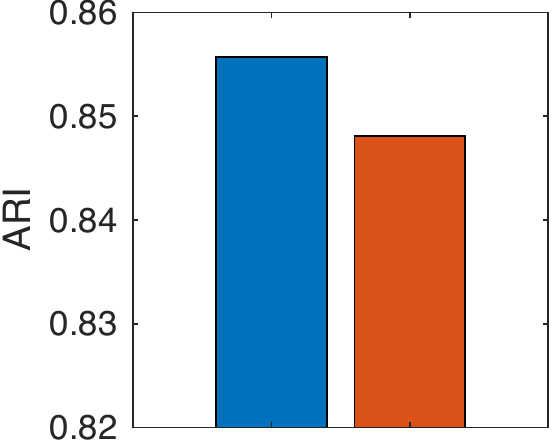}}}
		{\subfigure[ImageNet-Dogs]
			{\includegraphics[width=0.26\columnwidth]{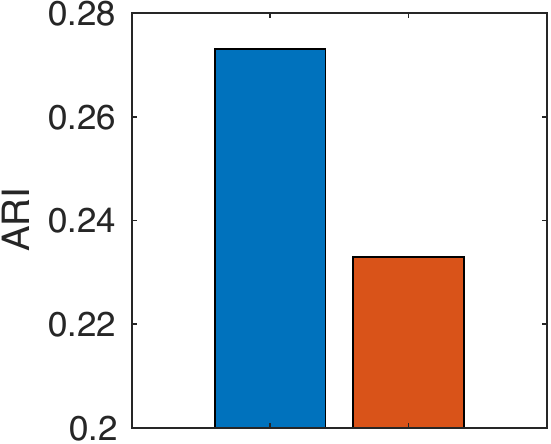}}}
		{\subfigure[Tiny-ImageNet]
			{\includegraphics[width=0.26\columnwidth]{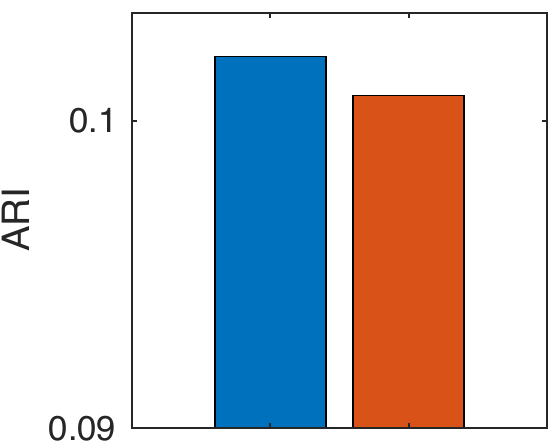}}}
		{\subfigure
			{\includegraphics[width=0.65\columnwidth]{figure/weighted_clusters/cmpWeightedClusterLegend}}}
		\caption{The ARI performance of DeepCluE with or without weighted clusters.}
		\label{fig:weighted_clusters_ari}
	\end{center}
\end{figure}

\begin{figure}[!t]
	\begin{center}
		{\subfigure[Fashion]
			{\includegraphics[width=0.26\columnwidth]{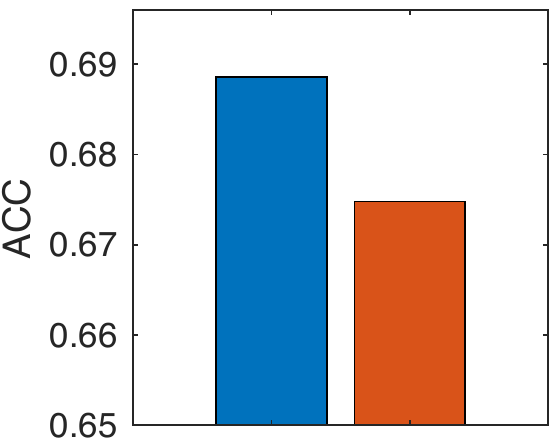}}}
		{\subfigure[CIFAR-10]
			{\includegraphics[width=0.26\columnwidth]{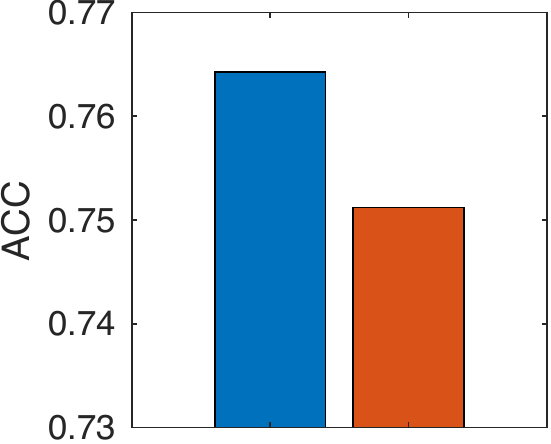}}}
		{\subfigure[CIFAR-100]
			{\includegraphics[width=0.26\columnwidth]{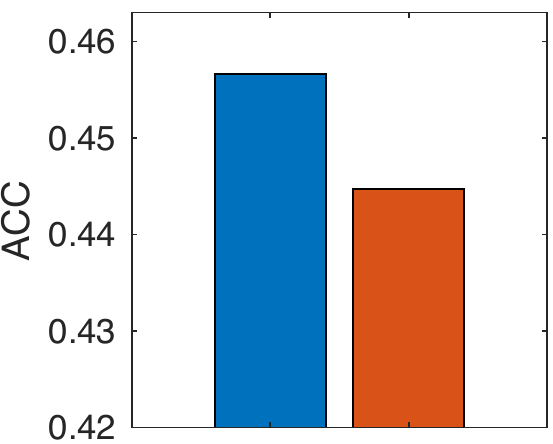}}}\\
		{\subfigure[ImageNet-10]
			{\includegraphics[width=0.26\columnwidth]{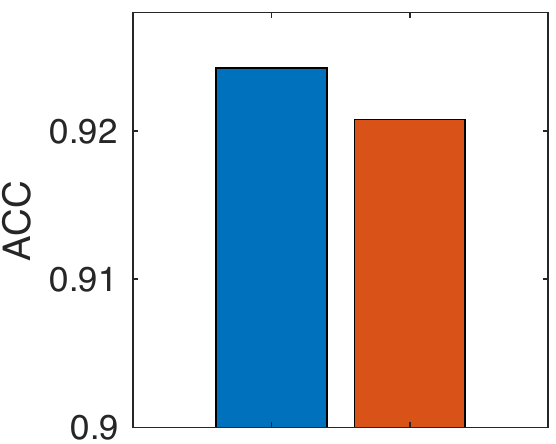}}}
		{\subfigure[ImageNet-Dogs]
			{\includegraphics[width=0.26\columnwidth]{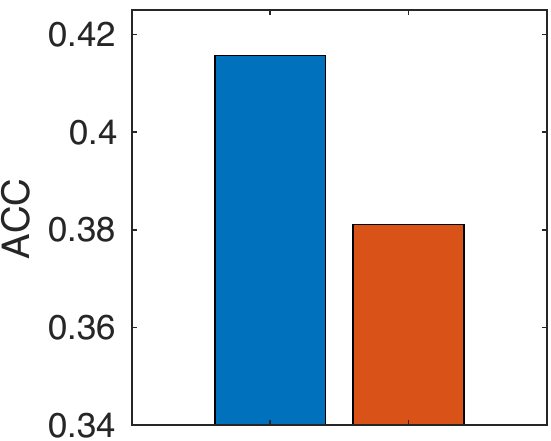}}}
		{\subfigure[Tiny-ImageNet]
			{\includegraphics[width=0.26\columnwidth]{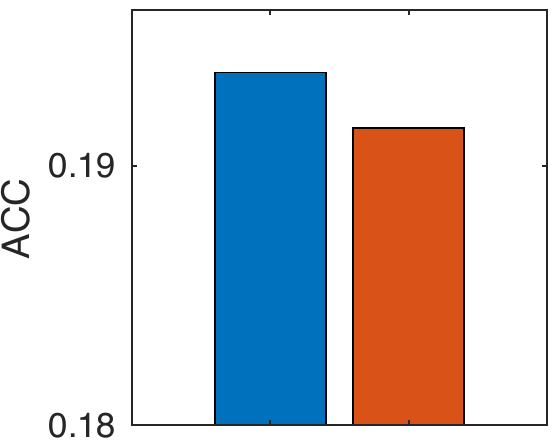}}}
		{\subfigure
			{\includegraphics[width=0.65\columnwidth]{figure/weighted_clusters/cmpWeightedClusterLegend}}}
		\caption{The ACC performance of DeepCluE with or without weighted clusters.}
		\label{fig:weighted_clusters_acc}
	\end{center}
\end{figure}

\subsection{Influence of Weighted Clusters}

In DeepCluE, a cluster-wise weighting strategy is incorporated to take the different reliability of multiple base clusterings (as well as the different clusters inside the same base clustering) into account. Note that the diversity is one of the key factors in ensemble clustering. As it is not required that the every base clustering has high reliability, the diversified base clusterings generated from multiple network layers is crucial for building a better consensus clustering result. In this section, we evaluate the influence of the weighted clusters in our DeepCluE method. Especially, the variant without weighted clusters can be achieved by simply setting the weights of all clusters to equally one. As shown in Figs.~\ref{fig:weighted_clusters_nmi}, \ref{fig:weighted_clusters_ari}, and \ref{fig:weighted_clusters_acc}, our DeepCluE method with weighted clusters consistently outperforms the variant without weighted clusters on the benchmark datasets, which confirm the substantial contribution of the cluster-wise weighting scheme in our DeepCluE method.

\subsection{Influence of Ensemble Size}
\label{sec:ensembel_size}

\begin{figure}[!t]
	\begin{center}
		{\subfigure[Fashion]
			{\includegraphics[width=0.3\columnwidth]{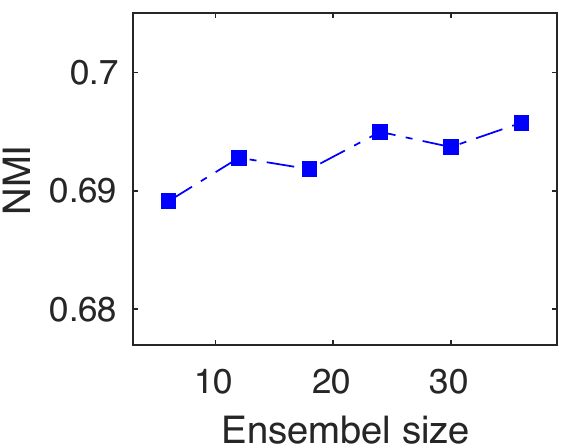}}}
		{\subfigure[CIFAR-10]
			{\includegraphics[width=0.3\columnwidth]{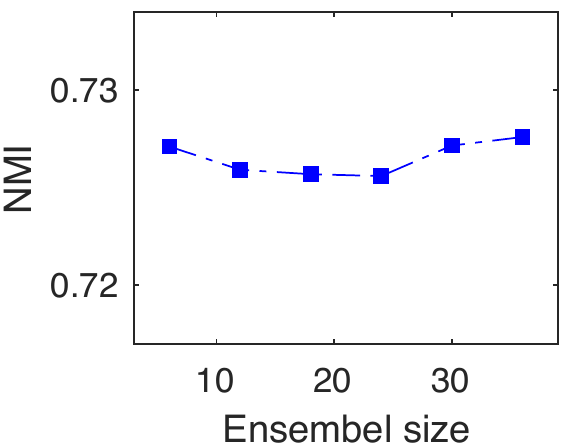}}}
		{\subfigure[CIFAR-100]
			{\includegraphics[width=0.3\columnwidth]{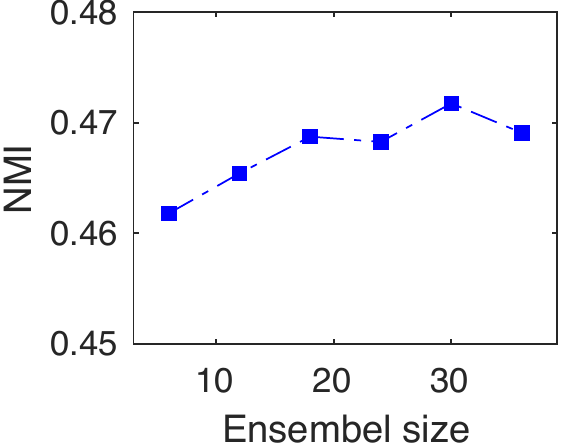}}}\\
		{\subfigure[ImageNet-10]
			{\includegraphics[width=0.3\columnwidth]{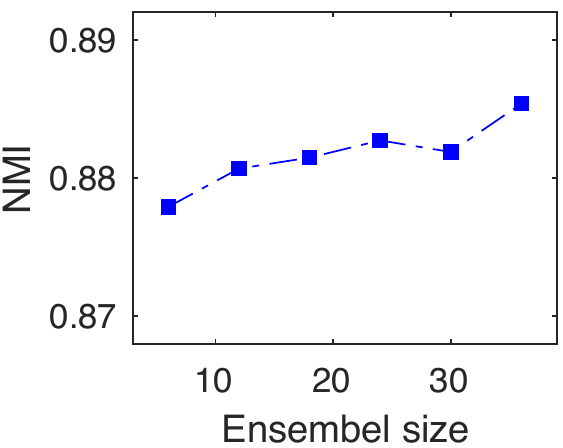}}}
		{\subfigure[ImageNet-Dogs]
			{\includegraphics[width=0.3\columnwidth]{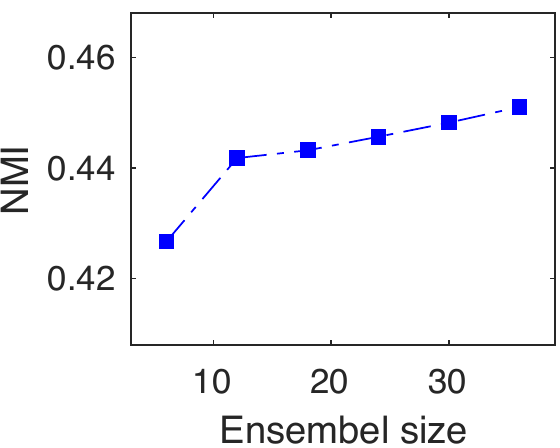}}}
		{\subfigure[Tiny-ImageNet]
			{\includegraphics[width=0.3\columnwidth]{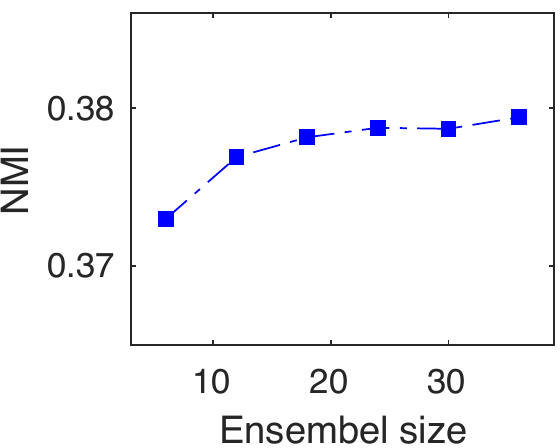}}}
		\caption{The NMI performance of DeepCluE with varying ensemble sizes.}
		\label{fig:ensemble_size_nmi}
	\end{center}
\end{figure}

\begin{figure}[!t]
	\begin{center}
		{\subfigure[Fashion]
			{\includegraphics[width=0.3\columnwidth]{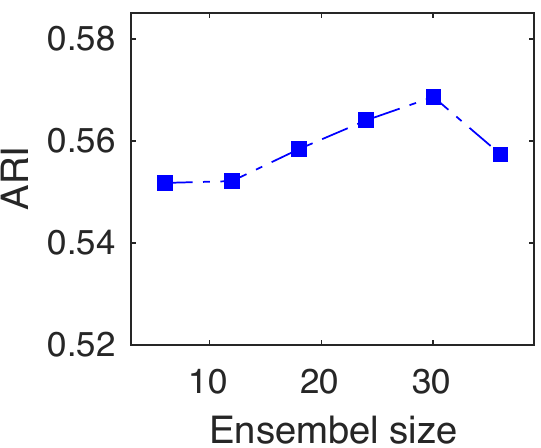}}}
		{\subfigure[CIFAR-10]
			{\includegraphics[width=0.3\columnwidth]{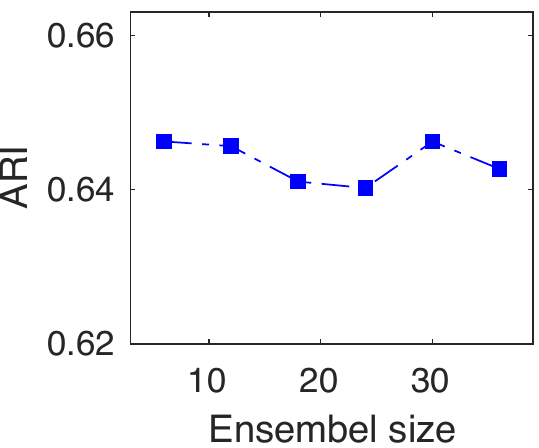}}}
		{\subfigure[CIFAR-100]
			{\includegraphics[width=0.3\columnwidth]{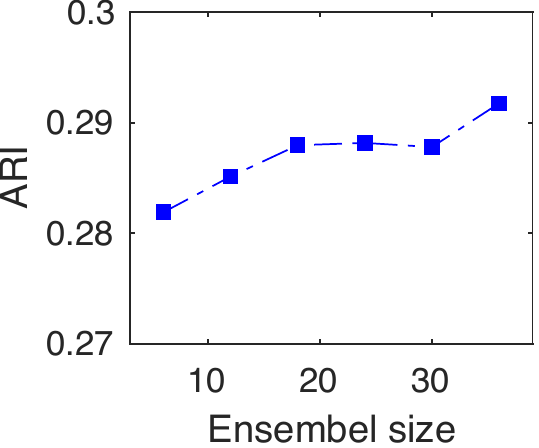}}}\\
		{\subfigure[ImageNet-10]
			{\includegraphics[width=0.3\columnwidth]{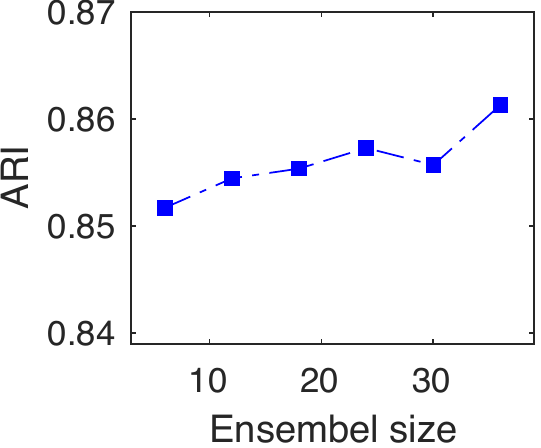}}}
		{\subfigure[ImageNet-Dogs]
			{\includegraphics[width=0.3\columnwidth]{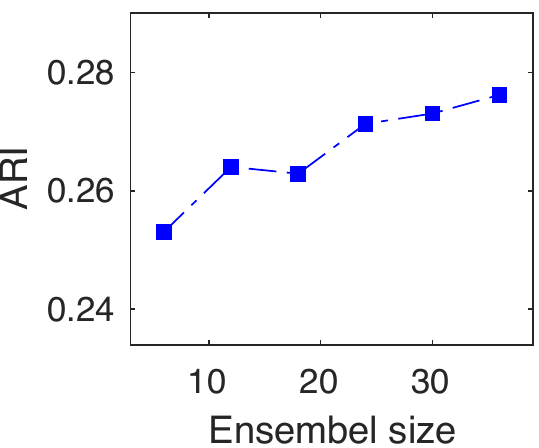}}}
		{\subfigure[Tiny-ImageNet]
			{\includegraphics[width=0.3\columnwidth]{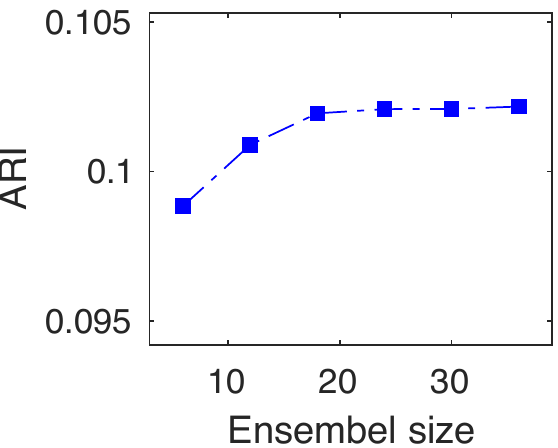}}}
		\caption{The ARI performance of DeepCluE with varying ensemble sizes.}
		\label{fig:ensemble_size_ari}
	\end{center}
\end{figure}

\begin{figure}[!t]
	\begin{center}
		{\subfigure[Fashion]
			{\includegraphics[width=0.3\columnwidth]{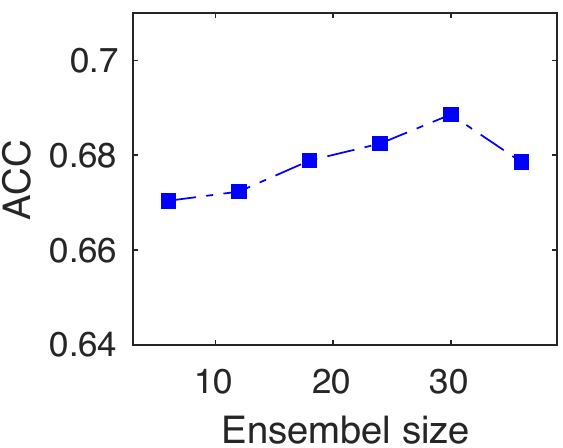}}}
		{\subfigure[CIFAR-10]
			{\includegraphics[width=0.3\columnwidth]{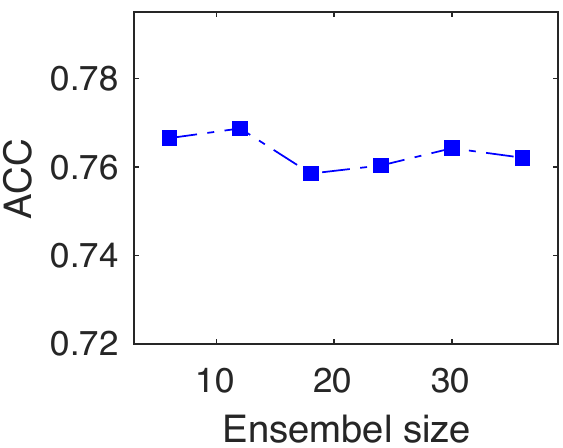}}}
		{\subfigure[CIFAR-100]
			{\includegraphics[width=0.3\columnwidth]{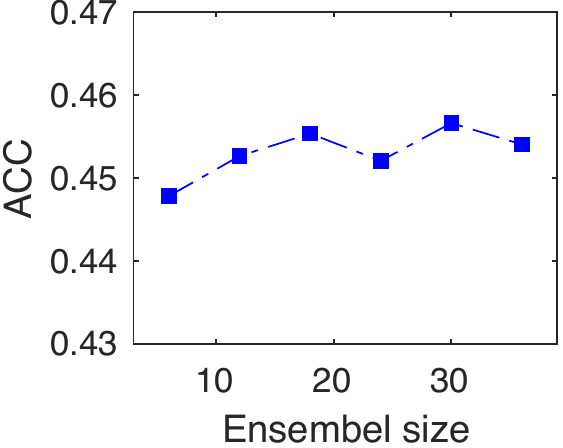}}}\\
		{\subfigure[ImageNet-10]
			{\includegraphics[width=0.3\columnwidth]{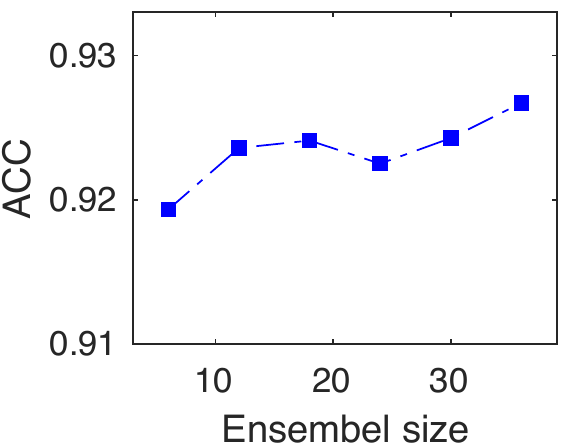}}}
		{\subfigure[ImageNet-Dogs]
			{\includegraphics[width=0.3\columnwidth]{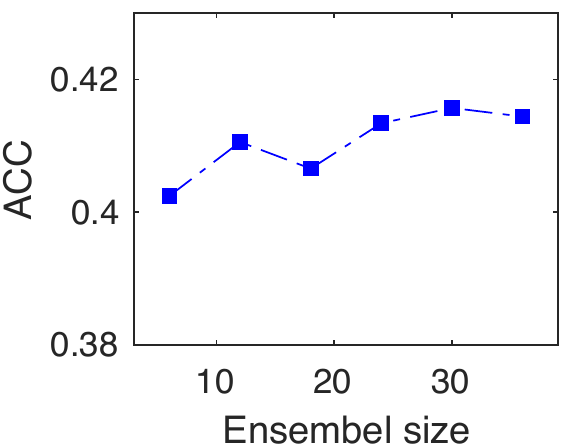}}}
		{\subfigure[Tiny-ImageNet]
			{\includegraphics[width=0.3\columnwidth]{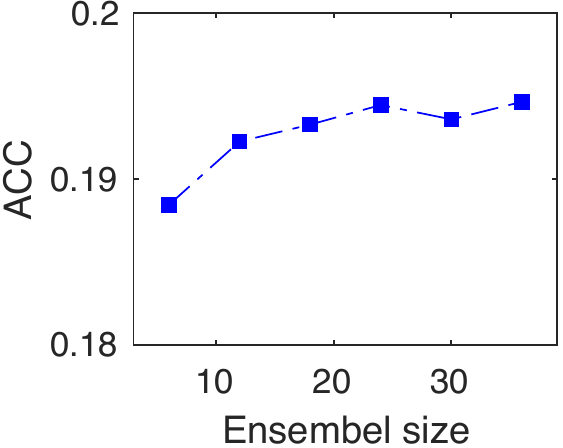}}}
		\caption{The ACC performance of DeepCluE with varying ensemble sizes.}
		\label{fig:ensemble_size_acc}
	\end{center}
\end{figure}

In the proposed DeepCluE method, multiple layers of representations are jointly utilized in an ensemble clustering manner. In this section, we evaluate the influence of the number of base clusterings $M$ in our framework. We illustrate the NMI, ARI, and ACC scores of DeepCluE with different ensemble sizes in Figs.~\ref{fig:ensemble_size_nmi}, \ref{fig:ensemble_size_ari}, and \ref{fig:ensemble_size_acc}, respectively. Note that on each layer, we produce $M'$ base clusterings. Thus a total of $M=\lambda M'$ base clusterings are generated on $\lambda$ layers of representations. When $M'$ goes from 1 to 6, with $\lambda=6$ layers of representations utilized, the total number of base clusterings $M$ grows from 6 to 36. As can be observed in the performance curves, DeepCluE is able to yield stably high-quality clustering results with varying number of base clusterings on the benchmark datasets. Typically, a relative larger ensemble size can often be beneficial to the clustering performance.

\section{Conclusion and Future Work}
\label{sec:conclusion}

In this paper, we propose a novel deep image clustering approach termed DeepCluE, which bridges the gap between deep clustering and ensemble clustering. Different from previous deep clustering approaches that mostly utilize a single layer of representation to construct the final clustering, the DeepCluE approach jointly exploits multiple layers of feature representations in the deep neural network by means of an ensemble clustering process. Specifically, a weight-sharing convolutional neural network is first trained with two separate projectors, i.e., the instance projector and the cluster projector. Then multi-layer representations are extracted from the trained network for producing a set of diversified base clusterings via the efficient U-SPEC algorithm. Further, an entropy-based criterion is adopted to evaluate and weight the clusters in multiple base clusterings, through which a weighted-cluster bipartite graph can further be formulated and partitioned for the final clustering. Extensive experimental results on six well-known image datasets have demonstrated the superiority of the proposed DeepCluE approach over the state-of-the-art deep clustering approaches. 

In terms of the limitations and future directions, in this paper, we mainly focus on the deep clustering task for image data. In the future work, our framework can be extended to enforce the deep ensemble clustering for more types of data, such the document data and the time series data. Besides this, another limitation to the current work is that the extraction of the feature representations of multiple layers in the deep network may involve large storage space (depending on the selected backbone network), which gives rise to another potential research direction in the future extensions.

\ifCLASSOPTIONcompsoc
\section*{Acknowledgments}
\else
\section*{Acknowledgment}
\fi

This project was supported by the NSFC (61976097, 62276277  \& 62076258), the Natural Science Foundation of Guangdong Province (2021A1515012203), and the Science and Technology Program of Guangzhou, China (202201010314).

\ifCLASSOPTIONcaptionsoff
\newpage
\fi

\bibliographystyle{IEEEtran}
\bibliography{Refs}

\begin{thebibliography}{10}
\providecommand{\url}[1]{#1}
\csname url@samestyle\endcsname
\providecommand{\newblock}{\relax}
\providecommand{\bibinfo}[2]{#2}
\providecommand{\BIBentrySTDinterwordspacing}{\spaceskip=0pt\relax}
\providecommand{\BIBentryALTinterwordstretchfactor}{4}
\providecommand{\BIBentryALTinterwordspacing}{\spaceskip=\fontdimen2\font plus
\BIBentryALTinterwordstretchfactor\fontdimen3\font minus
  \fontdimen4\font\relax}
\providecommand{\BIBforeignlanguage}[2]{{%
\expandafter\ifx\csname l@#1\endcsname\relax
\typeout{** WARNING: IEEEtran.bst: No hyphenation pattern has been}%
\typeout{** loaded for the language `#1'. Using the pattern for}%
\typeout{** the default language instead.}%
\else
\language=\csname l@#1\endcsname
\fi
#2}}
\providecommand{\BIBdecl}{\relax}
\BIBdecl

\bibitem{jain2010data}
A.~K. Jain, ``Data clustering: 50 years beyond k-means,'' \emph{Pattern
  Recognition Letters}, vol.~31, no.~8, pp. 651--666, 2010.

\bibitem{yang2017towards}
B.~Yang, X.~Fu, N.~D. Sidiropoulos, and M.~Hong, ``Towards k-means-friendly
  spaces: Simultaneous deep learning and clustering,'' in \emph{Proc. of
  International Conference on Machine Learning (ICML)}, 2017, pp. 3861--3870.

\bibitem{xie2016unsupervised}
J.~Xie, R.~Girshick, and A.~Farhadi, ``Unsupervised deep embedding for
  clustering analysis,'' in \emph{Proc. of International Conference on Machine
  Learning (ICML)}, 2016, pp. 478--487.

\bibitem{yang2016joint}
J.~Yang, D.~Parikh, and D.~Batra, ``Joint unsupervised learning of deep
  representations and image clusters,'' in \emph{Proc. of IEEE Conference on
  Computer Vision and Pattern Recognition (CVPR)}, 2016, pp. 5147--5156.

\bibitem{guo2017improved}
X.~Guo, L.~Gao, X.~Liu, and J.~Yin, ``Improved deep embedded clustering with
  local structure preservation.'' in \emph{Proc. of International Joint
  Conference on Artificial Intelligence (IJCAI)}, 2017, pp. 1753--1759.

\bibitem{ghasedi2017deep}
K.~G. Dizaji, A.~Herandi, C.~Deng, W.~Cai, and H.~Huang, ``Deep clustering via
  joint convolutional autoencoder embedding and relative entropy
  minimization,'' in \emph{Proc. of IEEE International Conference on Computer
  Vision (ICCV)}, 2017, pp. 5736--5745.

\bibitem{caron2018deep}
M.~Caron, P.~Bojanowski, A.~Joulin, and M.~Douze, ``Deep clustering for
  unsupervised learning of visual features,'' in \emph{Proc. of European
  Conference on Computer Vision (ECCV)}, 2018, pp. 132--149.

\bibitem{ji2019invariant}
X.~Ji, J.~F. Henriques, and A.~Vedaldi, ``Invariant information clustering for
  unsupervised image classification and segmentation,'' in \emph{Proc. of IEEE
  International Conference on Computer Vision (ICCV)}, 2019, pp. 9865--9874.

\bibitem{Choudhury2021}
S.~J. Choudhury and N.~R. Pal, ``Deep and structure-preserving autoencoders for
  clustering data with missing information,'' \emph{IEEE Transactions on
  Emerging Topics in Computational Intelligence}, vol.~5, no.~4, pp. 639--650,
  2021.

\bibitem{huang2020deep}
J.~Huang, S.~Gong, and X.~Zhu, ``Deep semantic clustering by partition
  confidence maximisation,'' in \emph{Proc. of IEEE Conference on Computer
  Vision and Pattern Recognition (CVPR)}, 2020, pp. 8849--8858.

\bibitem{li2021contrastive}
Y.~Li, P.~Hu, Z.~Liu, D.~Peng, J.~T. Zhou, and X.~Peng, ``Contrastive
  clustering,'' in \emph{Proc. of AAAI Conference on Artificial Intelligence
  (AAAI)}, 2021.

\bibitem{van2020scan}
W.~Van~Gansbeke, S.~Vandenhende, S.~Georgoulis, M.~Proesmans, and L.~Van~Gool,
  ``{SCAN}: Learning to classify images without labels,'' in \emph{Proc. of
  European Conference on Computer Vision (ECCV)}, 2020, pp. 268--285.

\bibitem{dang2021nearest}
Z.~Dang, C.~Deng, X.~Yang, K.~Wei, and H.~Huang, ``Nearest neighbor matching
  for deep clustering,'' in \emph{Proc. of IEEE Conference on Computer Vision
  and Pattern Recognition (CVPR)}, 2021, pp. 13\,693--13\,702.

\bibitem{Deng2023}
X.~Deng, D.~Huang, D.-H. Chen, C.-D. Wang, and J.-H. Lai, ``Strongly augmented
  contrastive clustering,'' \emph{Pattern Recognition}, vol. 139, p. 109470,
  2023.

\bibitem{xu19_ijcai}
C.~Xu, Z.~Guan, W.~Zhao, H.~Wu, Y.~Niu, and B.~Ling, ``Adversarial incomplete
  multi-view clustering,'' in \emph{Proc. of International Joint Conference on
  Artificial Intelligence (IJCAI)}, 2019, p. 3933–3939.

\bibitem{xu22_tcyb}
C.~Xu, H.~Liu, Z.~Guan, X.~Wu, J.~Tan, and B.~Ling, ``Adversarial incomplete
  multiview subspace clustering networks,'' \emph{IEEE Transactions on
  Cybernetics}, vol.~52, no.~10, pp. 10\,490--10\,503, 2022.

\bibitem{huang2015robust}
D.~Huang, J.-H. Lai, and C.-D. Wang, ``Robust ensemble clustering using
  probability trajectories,'' \emph{IEEE Transactions on Knowledge and Data
  Engineering}, vol.~28, no.~5, pp. 1312--1326, 2015.

\bibitem{yu18_tkde}
Z.~Yu, P.~{Luo}, J.~{Liu}, H.~{Wong}, J.~{You}, G.~{Han}, and J.~{Zhang},
  ``Semi-supervised ensemble clustering based on selected constraint
  projection,'' \emph{IEEE Transactions on Knowledge and Data Engineering},
  vol.~30, no.~12, pp. 2394--2407, 2018.

\bibitem{huang19_tkde}
D.~Huang, C.-D. {Wang}, J.-S. {Wu}, J.-H. {Lai}, and C.-K. {Kwoh},
  ``Ultra-scalable spectral clustering and ensemble clustering,'' \emph{IEEE
  Transactions on Knowledge and Data Engineering}, vol.~32, no.~6, pp.
  1212--1226, 2020.

\bibitem{huang23_tkde}
D.~Huang, C.-D. Wang, and J.-H. Lai, ``Fast multi-view clustering via
  ensembles: Towards scalability, superiority, and simplicity,'' \emph{IEEE
  Transactions on Knowledge and Data Engineering, in press}, 2023.

\bibitem{cai2009locality}
D.~Cai, X.~He, X.~Wang, H.~Bao, and J.~Han, ``Locality preserving nonnegative
  matrix factorization,'' in \emph{Proc. of International Joint Conference on
  Artificial Intelligence (IJCAI)}, 2009.

\bibitem{ng2002spectral}
A.~Y. Ng, M.~I. Jordan, and Y.~Weiss, ``On spectral clustering: Analysis and an
  algorithm,'' in \emph{Advances in Neural Information Processing Systems
  (NeurIPS)}, 2002, pp. 849--856.

\bibitem{yang22_tetci}
J.~Yang and C.-T. Lin, ``Multi-view adjacency-constrained hierarchical
  clustering,'' \emph{IEEE Transactions on Emerging Topics in Computational
  Intelligence}, vol.~7, no.~4, pp. 1126--1138, 2023.

\bibitem{Cai2023}
X.~Cai, D.~Huang, G.-Y. Zhang, and C.-D. Wang, ``Seeking commonness and
  inconsistencies: A jointly smoothed approach to multi-view subspace
  clustering,'' \emph{Information Fusion}, vol.~91, pp. 364--375, 2023.

\bibitem{fang23_tetci}
S.-G. Fang, D.~Huang, C.-D. Wang, and Y.~Tang, ``Joint multi-view unsupervised
  feature selection and graph learning,'' \emph{IEEE Transactions on Emerging
  Topics in Computational Intelligence}, 2023.

\bibitem{peng2021attention}
Z.~Peng, H.~Liu, Y.~Jia, and J.~Hou, ``Attention-driven graph clustering
  network,'' in \emph{Proc. of ACM International Conference on Multimedia (ACM
  MM)}, 2021, pp. 935--943.

\bibitem{bo2020structural}
D.~Bo, X.~Wang, C.~Shi, M.~Zhu, E.~Lu, and P.~Cui, ``Structural deep clustering
  network,'' in \emph{Proc. of The Web Conference (WWW)}, 2020, pp. 1400--1410.

\bibitem{chiang2019cluster}
W.-L. Chiang, X.~Liu, S.~Si, Y.~Li, S.~Bengio, and C.-J. Hsieh,
  ``Cluster-{GCN}: An efficient algorithm for training deep and large graph
  convolutional networks,'' in \emph{Proc. of ACM SIGKDD International
  Conference on Knowledge Discovery and Data Mining}, 2019, pp. 257--266.

\bibitem{chen2020simple}
T.~Chen, S.~Kornblith, M.~Norouzi, and G.~Hinton, ``A simple framework for
  contrastive learning of visual representations,'' in \emph{Proc. of
  International Conference on Machine Learning (ICML)}, 2020, pp. 1597--1607.

\bibitem{chen2020improved}
X.~Chen, H.~Fan, R.~Girshick, and K.~He, ``Improved baselines with momentum
  contrastive learning,'' \emph{arXiv preprint arXiv:2003.04297}, 2020.

\bibitem{caron2020unsupervised}
M.~Caron, I.~Misra, J.~Mairal, P.~Goyal, P.~Bojanowski, and A.~Joulin,
  ``Unsupervised learning of visual features by contrasting cluster
  assignments,'' \emph{Advances in Neural Information Processing Systems
  (NeurIPS)}, vol.~33, pp. 9912--9924, 2020.

\bibitem{grill2020bootstrap}
J.-B. Grill, F.~Strub, F.~Altch{\'e}, C.~Tallec, P.~Richemond, E.~Buchatskaya,
  C.~Doersch, B.~Avila~Pires, Z.~Guo, M.~Gheshlaghi~Azar \emph{et~al.},
  ``Bootstrap your own latent-a new approach to self-supervised learning,''
  \emph{Advances in Neural Information Processing Systems (NeurIPS)}, vol.~33,
  pp. 21\,271--21\,284, 2020.

\bibitem{he2016deep}
K.~He, X.~Zhang, S.~Ren, and J.~Sun, ``Deep residual learning for image
  recognition,'' in \emph{Proc. of IEEE Conference on Computer Vision and
  Pattern Recognition (CVPR)}, 2016, pp. 770--778.

\bibitem{fred2005combining}
A.~L. Fred and A.~K. Jain, ``Combining multiple clusterings using evidence
  accumulation,'' \emph{IEEE Transactions on Pattern Analysis and Machine
  Intelligence}, vol.~27, no.~6, pp. 835--850, 2005.

\bibitem{iam2011link}
N.~Iam-On, T.~Boongoen, S.~Garrett, and C.~Price, ``A link-based approach to
  the cluster ensemble problem,'' \emph{IEEE Transactions on Pattern Analysis
  and Machine Intelligence}, vol.~33, no.~12, pp. 2396--2409, 2011.

\bibitem{yi2012robust}
J.~Yi, T.~Yang, R.~Jin, A.~K. Jain, and M.~Mahdavi, ``Robust ensemble
  clustering by matrix completion,'' in \emph{Proc. of IEEE International
  Conference on Data Mining (ICDM)}, 2012, pp. 1176--1181.

\bibitem{huang2016ensemble}
D.~Huang, J.~Lai, and C.-D. Wang, ``Ensemble clustering using factor graph,''
  \emph{Pattern Recognition}, vol.~50, pp. 131--142, 2016.

\bibitem{huang2017locally}
D.~Huang, C.-D. Wang, and J.-H. Lai, ``Locally weighted ensemble clustering,''
  \emph{IEEE Transactions on Cybernetics}, vol.~48, no.~5, pp. 1460--1473,
  2017.

\bibitem{huang2018enhanced}
D.~Huang, C.-D. Wang, H.~Peng, J.~Lai, and C.-K. Kwoh, ``Enhanced ensemble
  clustering via fast propagation of cluster-wise similarities,'' \emph{IEEE
  Transactions on Systems, Man, and Cybernetics: Systems}, vol.~51, no.~1, pp.
  508--520, 2021.

\bibitem{huang21_tcyb}
D.~Huang, C.-D. Wang, J.-H. Lai, and C.-K. Kwoh, ``Toward multidiversified
  ensemble clustering of high-dimensional data: From subspaces to metrics and
  beyond,'' \emph{IEEE Transactions on Cybernetics}, vol.~52, no.~11, pp.
  12\,231--12\,244, 2022.

\bibitem{li2012segmentation}
Z.~Li, X.-M. Wu, and S.-F. Chang, ``Segmentation using superpixels: A bipartite
  graph partitioning approach,'' in \emph{Proc. of IEEE Conference on Computer
  Vision and Pattern Recognition (CVPR)}, 2012, pp. 789--796.

\bibitem{xiao2017fashion}
H.~Xiao, K.~Rasul, and R.~Vollgraf, ``{Fashion-MNIST}: A novel image dataset
  for benchmarking machine learning algorithms,'' \emph{arXiv preprint
  arXiv:1708.07747}, 2017.

\bibitem{krizhevsky2009learning}
A.~Krizhevsky, G.~Hinton \emph{et~al.}, ``Learning multiple layers of features
  from tiny images,'' 2009.

\bibitem{deng2009imagenet}
J.~Deng, W.~Dong, R.~Socher, L.-J. Li, K.~Li, and L.~Fei-Fei, ``{ImageNet}: A
  large-scale hierarchical image database,'' in \emph{Proc. of IEEE Conference
  on Computer Vision and Pattern Recognition (CVPR)}, 2009, pp. 248--255.

\bibitem{Liang2022}
Y.~Liang, D.~Huang, C.-D. Wang, and P.~S. Yu, ``Multi-view graph learning by
  joint modeling of consistency and inconsistency,'' \emph{IEEE Transactions on
  Neural Networks and Learning Systems}, pp. 1--15, 2022.

\bibitem{Fang2023}
S.-G. Fang, D.~Huang, X.-S. Cai, C.-D. Wang, C.~He, and Y.~Tang, ``Efficient
  multi-view clustering via unified and discrete bipartite graph learning,''
  \emph{IEEE Transactions on Neural Networks and Learning Systems}, 2023.

\bibitem{bengio2006greedy}
Y.~Bengio, P.~Lamblin, D.~Popovici, and H.~Larochelle, ``Greedy layer-wise
  training of deep networks,'' \emph{Advances in Neural Information Processing
  Systems (NeurIPS)}, vol.~19, 2006.

\bibitem{vincent2010stacked}
P.~Vincent, H.~Larochelle, I.~Lajoie, Y.~Bengio, P.-A. Manzagol, and L.~Bottou,
  ``Stacked denoising autoencoders: Learning useful representations in a deep
  network with a local denoising criterion.'' \emph{Journal of Machine Learning
  Research}, vol.~11, no.~12, 2010.

\bibitem{radford2015unsupervised}
A.~Radford, L.~Metz, and S.~Chintala, ``Unsupervised representation learning
  with deep convolutional generative adversarial networks,'' \emph{arXiv
  preprint arXiv:1511.06434}, 2015.

\bibitem{zeiler2010deconvolutional}
M.~D. Zeiler, D.~Krishnan, G.~W. Taylor, and R.~Fergus, ``Deconvolutional
  networks,'' in \emph{Proc. of IEEE Conference on Computer Vision and Pattern
  Recognition (CVPR)}, 2010, pp. 2528--2535.

\bibitem{kingma2013auto}
D.~P. Kingma and M.~Welling, ``Auto-encoding variational {B}ayes,'' \emph{arXiv
  preprint arXiv:1312.6114}, 2013.

\bibitem{chang2017deep}
J.~Chang, L.~Wang, G.~Meng, S.~Xiang, and C.~Pan, ``Deep adaptive image
  clustering,'' in \emph{Proc. of IEEE International Conference on Computer
  Vision (ICCV)}, 2017, pp. 5879--5887.

\bibitem{wu2019deep}
J.~Wu, K.~Long, F.~Wang, C.~Qian, C.~Li, Z.~Lin, and H.~Zha, ``Deep
  comprehensive correlation mining for image clustering,'' in \emph{Proc. of
  IEEE Conference on Computer Vision and Pattern Recognition (CVPR)}, 2019, pp.
  8150--8159.

\bibitem{wang2021unsupervised}
X.~Wang, Z.~Liu, and S.~X. Yu, ``Unsupervised feature learning by cross-level
  instance-group discrimination,'' in \emph{Proc. of IEEE Conference on
  Computer Vision and Pattern Recognition (CVPR)}, 2021, pp. 12\,586--12\,595.

\bibitem{guo2022hcsc}
Y.~Guo, M.~Xu, J.~Li, B.~Ni, X.~Zhu, Z.~Sun, and Y.~Xu, ``{HCSC}: hierarchical
  contrastive selective coding,'' in \emph{Proc. of IEEE Conference on Computer
  Vision and Pattern Recognition (CVPR)}, 2022, pp. 9706--9715.

\bibitem{kingma2014adam}
D.~P. Kingma and J.~Ba, ``Adam: A method for stochastic optimization,''
  \emph{arXiv preprint arXiv:1412.6980}, 2014.

\end{thebibliography}

\end{document}